\documentclass[letterpaper,journal]{IEEEtran}
\usepackage{amsmath,amsfonts}
\usepackage{algorithmic}
\usepackage{algorithm}
\usepackage{array}
\usepackage[caption=false,font=footnotesize]{subfig}
\usepackage{textcomp}
\usepackage{stfloats}
\usepackage{url}
\usepackage{verbatim}
\usepackage{graphicx}
\usepackage{cite}
\usepackage{tikz}
\usepackage{threeparttable}
\usepackage{booktabs}
\usepackage{hyperref}
\usepackage{multirow}
\usepackage{tabularx}
\usepackage{tcolorbox}
\usepackage[table]{xcolor}
\begin{document}

\title{An Information-Theoretic Framework for Robust Large Language Model Editing}

\author{
\textbf{Qizhou Chen$^{1,2}$,  Chengyu Wang$^{2*}$, Taolin Zhang$^{3}$, Xiaofeng He$^{1*}\thanks{
*Corresponding authors: Prof. Xiaofeng He (hexf@cs.ecnu.edu.cn); Dr. Chengyu Wang (chengyu.wcy@alibaba-inc.com)
}$
}\\
$^{1}$East China Normal University, Shanghai, China 
$^{2}$Alibaba Group, Hangzhou, China\\
$^{3}$Hefei University of Technology, Hefei, China
}

\maketitle

\begin{abstract}

Large Language Models (LLMs) have become indispensable tools in science, technology, and society, enabling transformative advances across diverse fields. However, errors or outdated information within these models can undermine their accuracy and restrict their safe deployment. 
Developing efficient strategies for updating model knowledge without the expense and disruption of full retraining remains a critical challenge. Current model editing techniques frequently struggle to generalize corrections beyond narrow domains, leading to unintended consequences and limiting their practical impact.
Here, we introduce a novel framework for editing LLMs, grounded in information bottleneck theory. This approach precisely compresses and isolates the essential information required for generalizable knowledge correction while minimizing disruption to unrelated model behaviors.
Building upon this foundation, we present the Information Bottleneck Knowledge Editor (IBKE), which leverages compact latent representations to guide gradient-based updates, enabling robust and broadly applicable model editing.
We validate IBKE's effectiveness across multiple LLM architectures and standard benchmark tasks, demonstrating state-of-the-art accuracy and improved generality and specificity of edits. These findings establish a theoretically principled and practical paradigm for open-domain knowledge editing, advancing the utility and trustworthiness of LLMs in real-world applications.

\end{abstract}

\section{Introduction}

Large language models (LLMs)~\cite{DBLP:journals/corr/abs-2302-13971,DBLP:journals/fi/RoumeliotisT23,DBLP:conf/iclr/ZengLDWL0YXZXTM23} have revolutionized artificial intelligence, rapidly transforming fields such as finance~\cite{DBLP:journals/bigdatasociety/LuitseD21, DBLP:conf/acl/Li0L0L24, DBLP:conf/coling/LiuZJZLZZLC25}, medicine~\cite{DBLP:journals/artmed/NerellaBZCSBSSSBKR24, DBLP:journals/mlc/ZhengGCQLY25, DBLP:journals/npjdm/WuQLGLZWX25, zhou2025large}, and education~\cite{DBLP:journals/bjet/YanSZLMCLJG24, DBLP:journals/tvcg/GaoLSLYLSZWC25, DBLP:conf/sigcse/RaihanSSZ25}. 
Their remarkable ability to process and generate human-like text has made them indispensable across a wide range of applications. 
However, as LLMs are increasingly employed in complex, dynamic environments, inherited outdated or erroneous information from training data presents significant challenges~\cite{ZJUEditSurvey2023, DBLP:journals/csur/WangZLZCL25, DBLP:journals/corr/abs-2401-01286}, particularly in high-risk and knowledge-intensive sectors.
Efficiently updating the internal knowledge of LLMs without complete retraining has therefore become a crucial research priority. Model editing techniques make it possible to introduce targeted corrections, enabling models to integrate new knowledge or rectify errors while reducing computational demands and mitigating the risk of catastrophic forgetting~\cite{DBLP:conf/iclr/MitchellLBFM22, DBLP:conf/acl/HuangCWYLSYS24}.

\begin{figure*}[!t]
    \centering
    \includegraphics[width=1\textwidth]{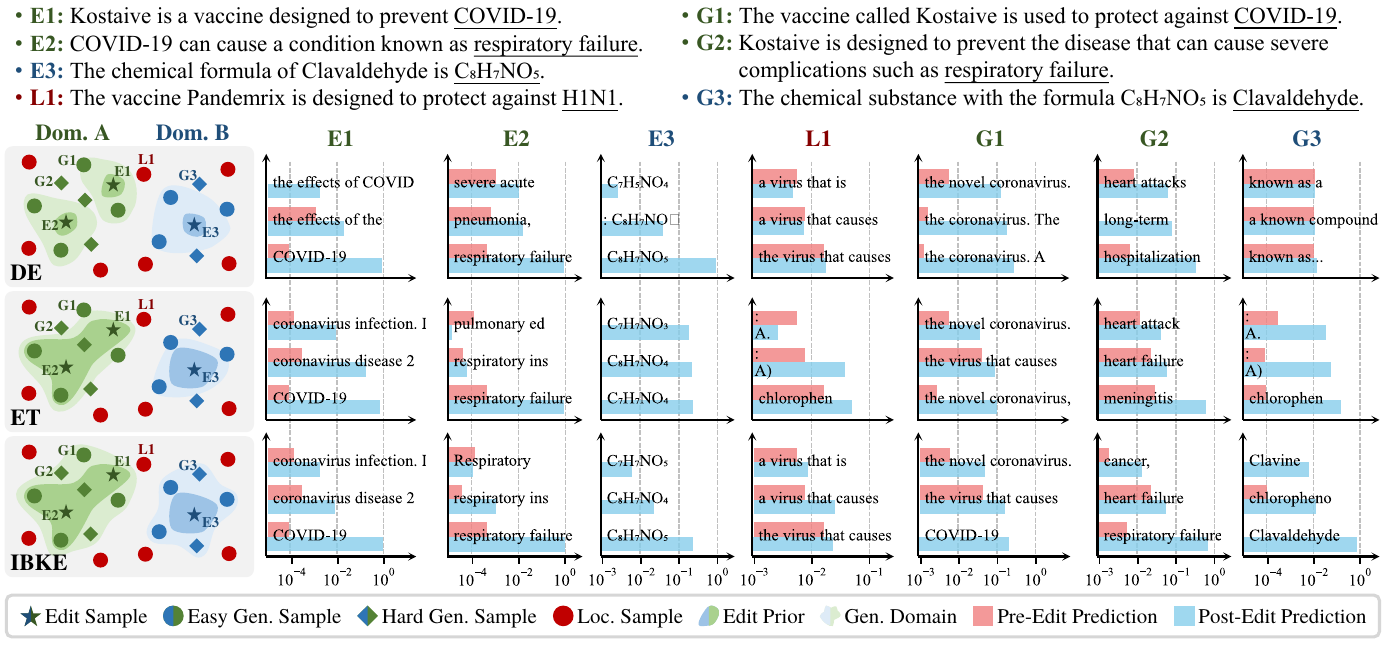}
\caption{
Comparison of two model editing paradigms with IBKE.
Direct Editing (DE) is based solely on the edit sample as its prior, while Edit Training (ET) leverages richer priors; when combined with the IB, ET achieves improved generalization.
The bar chart presents the top three prediction probabilities of the model after editing (using ``E1'', ``E2'', and ``E3'') for each corresponding sample.
Results for the three editing paradigms are shown as follows: ROME (for DE), IBKE without IB, and IBKE with IB, all implemented using the Qwen3-1.7B backbone model.
For illustration, Medicine (Domain A) and Chemistry (Domain B) are featured as example edit domains, with training performed on the Medicine domain.
An effective model editor should correct the model’s responses to generalization (Gen.) samples while preserving accuracy for locality (Loc.) samples.
}
\label{fig_three_edit_paradm}
\end{figure*}

Recent investigations into the explainability of knowledge localization within transformer models have spurred significant advances in knowledge editing~\cite{DBLP:conf/acl/DaiDHSCW22, VisEdit,ROME}. Contemporary approaches typically focus on identifying and directly modifying localized, knowledge-sensitive regions in the network, a process termed ``locate-then-edit'' (L\&E)~\cite{ROME, MEMIT, DBLP:conf/aaai/Li0SYMY24, AlphaEdit}. 
Alternatively, methods such as transformer patching~\cite{T-Patcher} bypass localization by optimizing a new neuron for each edit.
Despite their promise, both paradigms face notable limitations: optimizing a single piece of knowledge can often result in overfitting, leading to updates that are highly confident yet narrow and may fail to generalize to related queries or concepts~\cite{EVOKE}. 
As a consequence, performance on complex questions, especially those requiring multi-step reasoning or the handling of diverse inputs, is compromised~\cite{MQuAKE, UniEdit}. 
As illustrated in Figure~\ref{fig_three_edit_paradm}, reliance on a single edit prior restricts the domain of edit generality, making it simple and difficult to control.

Recent work has sought to enhance the edit prior by incorporating hypernetworks with edit training~\cite{KnowledgeEditor, SERAC, MEND, RECIPE, MALMEN}.
Meanwhile, benchmarks~\cite{MQuAKE, UniEdit} have been developed to broaden the scope of edit domains and increase diversity in edit generalization.
Nonetheless, overfitting continues to pose a challenge: model editors often excel on familiar data but struggle with out-of-domain knowledge, limiting their practical utility.
As shown in Figure~\ref{fig_three_edit_paradm}, edit training-based methods typically perform well in Domain~A, where the prior is dominant, but their performance diminishes in Domain~B.

To overcome these challenges, we propose a model editing framework grounded in the information bottleneck (IB) principle~\cite{DBLP:journals/corr/physics-0004057, DBLP:conf/iclr/AlemiFD017, DBLP:journals/pami/HuLYY24}. By constraining the flow of information during knowledge updates, the IB approach enables editors to extract and preserve features most relevant for generalization while omitting redundant or irrelevant details.
Conceptually, our method formulates model editing as a two-stage process: first, the edit request is transformed into a latent representation capturing the essential information; second, this representation guides a targeted intervention. 
The intervention may take various forms, such as parameter adaptation or the integration of auxiliary modules to steer model responses.
Our framework enforces three principles: minimizing transferred information to encourage compression, ensuring sufficiency for generalization so that edits propagate appropriately, and maintaining independence from unrelated knowledge to preserve locality.
Together, these constraints promote concise and impactful updates that generalize well beyond a single corrected instance.
As depicted in Figure~\ref{fig_three_edit_paradm}, applying the IB principle significantly expands the domain of edit generality, even when the editing prior remains fixed.

Building on these theoretical foundations, we present the Information Bottleneck Knowledge Editor (IBKE).
In its first stage, IBKE leverages first-order gradients with respect to the edit request~\cite{MEND, MALMEN}, distilling these signals into a compact latent space that forms the information bottleneck. In the second stage, the resulting edit representation is used to calibrate gradients and rescale update strengths across different tokens. These enhanced gradients are then applied to update the model weights.
Through edit training, IBKE incorporates training priors into the original gradients, thereby extending the domain of edit generality beyond specific instances.

We validate the effectiveness of IBKE in extensive experiments across four knowledge editing datasets: ZSRE~\cite{ZSRE}, CounterFact~\cite{ROME}, MQuAKE~\cite{MQuAKE}, and UniEdit~\cite{UniEdit}. These evaluations span multiple large language model architectures, including GPT2-XL (1.5B), GPT-J (6B), Qwen3-1.5B, and Qwen3-8B. Our results show substantial improvements over existing methods, positioning IBKE as a robust and generalizable solution for open-domain model editing.

\section{Results}

\subsection{Experimental Settings}
\noindent\textbf{LLM Backbones:} We evaluated language model backbones with varying sizes and architectures, including GPT2-XL (1.5B)\footnote{\url{https://huggingface.co/openai-community/gpt2-xl}}, GPT-J (6B)\footnote{\url{https://huggingface.co/EleutherAI/gpt-j-6b}}, Qwen3-1.7B\footnote{\url{https://huggingface.co/Qwen/Qwen3-1.7B}}, and Qwen3-8B\footnote{\url{https://huggingface.co/Qwen/Qwen3-8B}}. 
All experiments employed full precision (FP32) for GPT2-XL and Qwen3-1.7B, and half precision (FP16) for GPT-J and Qwen3-8B.

\begin{table}[!t]
\caption{
Statistics of each benchmark, where CF, MH, OD, and CC denote Counterfactual, Multi-hop, Open-Domain, and Combinations of diverse criteria, respectively.}
    \begin{center}
    \setlength{\tabcolsep}{1pt}
    \renewcommand{\arraystretch}{1}
    \begin{threeparttable}
    \begin{tabularx}{31em}{
    >{
    \centering\arraybackslash}m{6.8em}
    |>{\centering\arraybackslash}m{5em}
    |>{\centering\arraybackslash}m{4.5em}
    |*{1}{>{\centering\arraybackslash}X}
    |*{1}{>{\centering\arraybackslash}X}
    |*{1}{>{\centering\arraybackslash}X}
    |*{1}{>{\centering\arraybackslash}X}
    }
    \toprule
\textbf{Datasets} & \textbf{Train Size} & \textbf{Test Size} & \textbf{CF} & \textbf{MH} & \textbf{OD} & \textbf{CC} \\
 \cmidrule(l{2pt}r{2pt}){1-1}
 \cmidrule(l{2pt}r{2pt}){2-2}
 \cmidrule(l{2pt}r{2pt}){3-3}
 \cmidrule(l{2pt}r{2pt}){4-4}
 \cmidrule(l{2pt}r{2pt}){5-5}
 \cmidrule(l{2pt}r{2pt}){6-6}
 \cmidrule(l{2pt}r{2pt}){7-7}
\textbf{ZSRE} & 163,196 & 19,086 &  &  &  &  \\
\textbf{CounterFact} & 10,000 & 10,000 & $\checkmark$ &  &  &  \\
\textbf{MQuAKE} & 2,036 & 10,182 & $\checkmark$ & $\checkmark$ &  &  \\
\textbf{UniEdit} & 207,419 & 103,723 &  & $\checkmark$ & $\checkmark$ & $\checkmark$\\
\bottomrule
    \end{tabularx}
    \end{threeparttable}
    \end{center}
    \label{tab_statistics_benchmarks}
\end{table}

\noindent\textbf{Baseline Editors:}
To comprehensively assess editing approaches, we categorized methods into two principal groups: Direct Editing (DE) and Edit Training (ET).  
Representative editors from both categories were selected to ensure coverage of diverse paradigms.  
DE methods include those that directly modify the model parameters, such as Fine-Tuning (FT)~\cite{ZJUEditSurvey2023}, ROME~\cite{ROME}, and AlphaEdit~\cite{AlphaEdit}, as well as techniques employing external parameters or modules, including T-Patcher~\cite{T-Patcher}, LEMoE~\cite{LEMOE}, and GRACE~\cite{GRACE}.
The ET category includes MEND~\cite{MEND}, which utilizes a multilayer perceptron (MLP)-based hypernetwork to adjust model parameters, and SERAC~\cite{SERAC}, which incorporates a lightweight counterfactual model to generate edit-related responses.
Baseline hyperparameters were set in accordance with those reported in~\cite{wang2024easyedit}.

\noindent\textbf{IBKE Settings:}
Building on prior research in knowledge localization, attribution analysis, and related editing techniques~\cite{ROME, MEMIT, wang2024easyedit}, 
we selected the output linear layer matrices from the feed-forward networks (FFNs) within editing-sensitive layers of each backbone model as the edit weights, denoted $\{W_i\}_{i=1}^n$. 
Specifically, we applied edits to layers 15, 16, and 17 in GPT2-XL; layers 6, 7, and 8 in GPT-J; and layers 18, 19, and 20 in both Qwen3 variants. 
Consistent with the approach described in~\cite{MEND}, we set the dimension of the hypernetwork module to $d_m = 1920$. 
Based on hyperparameter searching, the sequence length parameter was set to $l_m = 10$, and the IB regularization coefficient to $\beta = 0.1$.
For edit training, we used a learning rate of $1\times10^{-4}$ and a maximum batch size of 8, with batch size adjusted dynamically according to the interactions among edit samples.

\begin{table*}[!t]
\caption{
Overall editing performance on UniEdit, MQuAKE, CounterFact, and ZSRE. ``W/O'' indicates results for pre-edit LLMs. ``Rel.'', ``Gen.'', and ``Loc.'' denote reliability, generality, and locality, respectively. 
\textbf{DE} and \textbf{ET} refer to editor types based on Direct Editing and Edit Training, respectively. 
The training dataset for ET methods marked with an asterisk (*) is augmented with a small amount of additional data.
The best performance is highlighted in dark blue, and results within 3\% of the best are highlighted in light blue.
}

    \begin{center}
    \scriptsize
    \setlength{\tabcolsep}{1pt}
    \renewcommand{\arraystretch}{1}
    \begin{threeparttable}
    \begin{tabularx}{73.5em}{
    >{\centering\arraybackslash}m{5.3em}|
    >{\centering\arraybackslash}m{2.8em}|
    >{\centering\arraybackslash}m{5.3em}|
    *{3}{>{\centering\arraybackslash}X}
    >{\centering\arraybackslash}m{5.5em}|
    *{2}{>{\centering\arraybackslash}X}
    >{\centering\arraybackslash}m{5.5em}|
    *{3}{>{\centering\arraybackslash}X}
    >{\centering\arraybackslash}m{5.5em}|
    *{3}{>{\centering\arraybackslash}X}
    >{\centering\arraybackslash}m{5.5em}
    }

\toprule
 &  &  & \multicolumn{4}{c|}{\textbf{UniEdit}} & \multicolumn{3}{c|}{\textbf{MQuAKE}} & \multicolumn{4}{c|}{\textbf{CounterFact}} & \multicolumn{4}{c}{\textbf{ZSRE}} \\
\multirow{-2}{*}{\textbf{Backbones}} & \multirow{-2}{*}{\textbf{Type}} & \multirow{-2}{*}{\textbf{Editors}} & \textbf{Rel.} & \textbf{Gen.} & \textbf{Loc.} & \textbf{Avg.} & \textbf{Rel.} & \textbf{Gen.} & \textbf{Avg.} & \textbf{Rel.} & \textbf{Gen.} & \textbf{Loc.} & \textbf{Avg.} & \textbf{Rel.} & \textbf{Gen.} & \textbf{Loc.} & \textbf{Avg.} \\
\cmidrule(l{2pt}r{2pt}){1-1}
\cmidrule(l{2pt}r{2pt}){2-2}
 \cmidrule(l{2pt}r{2pt}){3-3}
 \cmidrule(l{2pt}r{2pt}){4-7}
 \cmidrule(l{2pt}r{2pt}){8-10}
 \cmidrule(l{2pt}r{2pt}){11-14}
 \cmidrule(l{2pt}r{2pt}){15-18}
 &  & \textbf{W/O} & 20.32 & 27.53 & \cellcolor[HTML]{5B9BD5}100.00 & 49.28$_{\pm0.02}$ & 32.42 & 25.10 & 28.76$_{\pm0.05}$ & 0.00 & 3.42 & \cellcolor[HTML]{5B9BD5}100.00 & 34.47$_{\pm0.01}$ & 19.26 & 27.41 & \cellcolor[HTML]{5B9BD5}100.00 & 48.89$_{\pm0.03}$ \\
 &  & \textbf{FT} & \cellcolor[HTML]{BDD7EE}99.10 & 60.43 & 89.68 & 83.07$_{\pm0.56}$ & \cellcolor[HTML]{BDD7EE}99.80 & 44.08 & 71.94$_{\pm0.23}$ & \cellcolor[HTML]{BDD7EE}99.25 & 58.34 & 77.56 & 78.38$_{\pm0.36}$ & \cellcolor[HTML]{BDD7EE}99.47 & \cellcolor[HTML]{BDD7EE}96.75 & \cellcolor[HTML]{BDD7EE}97.89 & \cellcolor[HTML]{BDD7EE}98.04$_{\pm0.17}$ \\
 &  & \textbf{ROME} & \cellcolor[HTML]{BDD7EE}96.76 & 58.16 & 96.76 & 83.90$_{\pm0.31}$ & 89.61 & 22.90 & 56.26$_{\pm0.45}$ & \cellcolor[HTML]{BDD7EE}99.35 & 46.71 & 95.91 & 80.66$_{\pm0.37}$ & \cellcolor[HTML]{BDD7EE}99.58 & 85.49 & \cellcolor[HTML]{BDD7EE}99.14 & 94.74$_{\pm0.20}$ \\
 &  & \textbf{T-Patcher} & 90.61 & 59.12 & 87.97 & 79.23$_{\pm0.31}$ & \cellcolor[HTML]{BDD7EE}97.95 & 34.62 & 66.28$_{\pm0.37}$ & \cellcolor[HTML]{BDD7EE}99.00 & 33.72 & 88.21 & 73.64$_{\pm0.13}$ & 94.42 & \cellcolor[HTML]{BDD7EE}96.86 & \cellcolor[HTML]{BDD7EE}98.35 & 96.54$_{\pm0.15}$ \\
 &  & \textbf{LEMoE} & 78.87 & 60.18 & 89.61 & 76.22$_{\pm0.24}$ & 72.68 & 41.80 & 57.24$_{\pm0.52}$ & \cellcolor[HTML]{5B9BD5}99.78 & 47.45 & 93.20 & 80.14$_{\pm0.03}$ & \cellcolor[HTML]{BDD7EE}97.50 & \cellcolor[HTML]{BDD7EE}96.44 & \cellcolor[HTML]{BDD7EE}99.03 & \cellcolor[HTML]{BDD7EE}97.66$_{\pm0.12}$ \\
 &  & \textbf{GRACE} & \cellcolor[HTML]{5B9BD5}99.64 & 46.58 & \cellcolor[HTML]{BDD7EE}98.75 & 81.66$_{\pm0.09}$ & \cellcolor[HTML]{5B9BD5}99.95 & 24.99 & 62.47$_{\pm0.04}$ & \cellcolor[HTML]{BDD7EE}99.57 & 3.77 & \cellcolor[HTML]{BDD7EE}98.59 & 67.31$_{\pm0.15}$ & \cellcolor[HTML]{BDD7EE}99.11 & 27.52 & \cellcolor[HTML]{BDD7EE}99.80 & 75.48$_{\pm0.13}$ \\
 & \multirow{-7}{*}{\textbf{DE}} & \textbf{AlphaEdit} & \cellcolor[HTML]{BDD7EE}97.01 & 59.00 & 96.67 & 84.23$_{\pm0.42}$ & 95.21 & 24.69 & 59.95$_{\pm0.40}$ & \cellcolor[HTML]{BDD7EE}99.12 & 62.58 & \cellcolor[HTML]{BDD7EE}98.58 & 86.76$_{\pm0.31}$ & \cellcolor[HTML]{BDD7EE}99.10 & 93.27 & \cellcolor[HTML]{BDD7EE}99.48 & \cellcolor[HTML]{BDD7EE}97.29$_{\pm0.25}$ \\
\cmidrule(l{2pt}r{2pt}){2-2}
 \cmidrule(l{2pt}r{2pt}){3-3}
 \cmidrule(l{2pt}r{2pt}){4-7}
 \cmidrule(l{2pt}r{2pt}){8-10}
 \cmidrule(l{2pt}r{2pt}){11-14}
 \cmidrule(l{2pt}r{2pt}){15-18}
 &  & \textbf{SERAC} & \cellcolor[HTML]{BDD7EE}98.59 & 78.39 & 87.36 & 88.11$_{\pm0.37}$ & 83.91 & 20.08 & 51.99$_{\pm0.11}$ & 77.17 & 28.94 & 82.12 & 62.75$_{\pm0.18}$ & 92.58 & 83.04 & \cellcolor[HTML]{BDD7EE}99.64 & 91.75$_{\pm0.18}$ \\
 &  & \textbf{MEND} & \cellcolor[HTML]{BDD7EE}97.40 & 68.92 & \cellcolor[HTML]{BDD7EE}97.99 & 88.10$_{\pm0.51}$ & \cellcolor[HTML]{BDD7EE}98.50 & 45.07 & 71.78$_{\pm0.14}$ & \cellcolor[HTML]{BDD7EE}97.65 & 43.30 & 83.43 & 74.79$_{\pm0.26}$ & \cellcolor[HTML]{BDD7EE}97.71 & 95.38 & 96.70 & 96.60$_{\pm0.17}$ \\
 &  & \textbf{IBKE} & \cellcolor[HTML]{BDD7EE}97.80 & \cellcolor[HTML]{BDD7EE}96.89 & \cellcolor[HTML]{BDD7EE}97.48 & \cellcolor[HTML]{BDD7EE}97.39$_{\pm0.30}$ & \cellcolor[HTML]{BDD7EE}97.34 & 56.09 & 76.72$_{\pm0.25}$ & 96.08 & 57.87 & 77.45 & 77.13$_{\pm0.18}$ & 95.79 & \cellcolor[HTML]{BDD7EE}97.69 & 96.36 & \cellcolor[HTML]{BDD7EE}96.61$_{\pm0.32}$ \\
 &  & \textbf{SERAC*} & \cellcolor[HTML]{BDD7EE}98.93 & 81.82 & 87.33 & 89.36$_{\pm0.16}$ & \cellcolor[HTML]{BDD7EE}99.38 & 47.71 & 73.54$_{\pm0.13}$ & \cellcolor[HTML]{BDD7EE}99.49 & 83.78 & 88.56 & \cellcolor[HTML]{BDD7EE}90.61$_{\pm0.16}$ & \cellcolor[HTML]{5B9BD5}99.92 & \cellcolor[HTML]{5B9BD5}99.02 & \cellcolor[HTML]{BDD7EE}99.85 & \cellcolor[HTML]{5B9BD5}99.60$_{\pm0.11}$ \\
 &  & \textbf{MEND*} & \cellcolor[HTML]{BDD7EE}97.22 & 69.64 & \cellcolor[HTML]{BDD7EE}97.19 & 88.02$_{\pm0.28}$ & 95.98 & 79.48 & 87.73$_{\pm0.03}$ & \cellcolor[HTML]{BDD7EE}97.73 & 76.69 & 89.18 & 87.87$_{\pm0.06}$ & \cellcolor[HTML]{BDD7EE}98.24 & \cellcolor[HTML]{BDD7EE}98.47 & \cellcolor[HTML]{BDD7EE}99.04 & \cellcolor[HTML]{BDD7EE}98.58$_{\pm0.06}$ \\
\multirow{-12.5}{*}{\textbf{\begin{tabular}[c]{@{}c@{}}GPT2-XL\\      (1.5B)\end{tabular}}} & \multirow{-6}{*}{\textbf{ET}} & \textbf{IBKE*} & \cellcolor[HTML]{BDD7EE}98.13 & \cellcolor[HTML]{5B9BD5}97.00 & \cellcolor[HTML]{BDD7EE}97.46 & \cellcolor[HTML]{5B9BD5}97.53$_{\pm0.17}$ & \cellcolor[HTML]{BDD7EE}98.72 & \cellcolor[HTML]{5B9BD5}92.31 & \cellcolor[HTML]{5B9BD5}95.51$_{\pm0.43}$ & \cellcolor[HTML]{BDD7EE}97.19 & \cellcolor[HTML]{5B9BD5}88.59 & 87.58 & \cellcolor[HTML]{5B9BD5}91.12$_{\pm0.22}$ & \cellcolor[HTML]{BDD7EE}97.08 & \cellcolor[HTML]{BDD7EE}97.83 & \cellcolor[HTML]{BDD7EE}99.24 & \cellcolor[HTML]{BDD7EE}98.05$_{\pm0.13}$ \\
\cmidrule(l{2pt}r{2pt}){1-1}
\cmidrule(l{2pt}r{2pt}){2-2}
 \cmidrule(l{2pt}r{2pt}){3-3}
 \cmidrule(l{2pt}r{2pt}){4-7}
 \cmidrule(l{2pt}r{2pt}){8-10}
 \cmidrule(l{2pt}r{2pt}){11-14}
 \cmidrule(l{2pt}r{2pt}){15-18}
 &  & \textbf{W/O} & 24.90 & 32.63 & \cellcolor[HTML]{5B9BD5}100.00 & 52.51$_{\pm0.05}$ & 35.24 & 26.43 & 30.84$_{\pm0.04}$ & 0.44 & 2.59 & \cellcolor[HTML]{5B9BD5}100.00 & 34.34$_{\pm0.05}$ & 21.56 & 29.92 & \cellcolor[HTML]{5B9BD5}100.00 & 50.50$_{\pm0.03}$ \\
 &  & \textbf{FT} & \cellcolor[HTML]{BDD7EE}99.60 & 64.75 & 90.57 & 84.98$_{\pm0.16}$ & \cellcolor[HTML]{BDD7EE}99.46 & 62.77 & 81.11$_{\pm0.44}$ & \cellcolor[HTML]{BDD7EE}99.90 & 62.94 & 92.22 & 85.02$_{\pm0.16}$ & \cellcolor[HTML]{5B9BD5}100.00 & \cellcolor[HTML]{BDD7EE}98.78 & \cellcolor[HTML]{BDD7EE}99.04 & \cellcolor[HTML]{BDD7EE}99.27$_{\pm0.11}$ \\
 &  & \textbf{ROME} & \cellcolor[HTML]{BDD7EE}97.26 & 64.68 & 95.81 & 85.92$_{\pm0.32}$ & 88.63 & 24.82 & 56.73$_{\pm0.35}$ & \cellcolor[HTML]{BDD7EE}99.20 & \cellcolor[HTML]{BDD7EE}83.82 & 93.91 & \cellcolor[HTML]{BDD7EE}92.31$_{\pm0.31}$ & \cellcolor[HTML]{BDD7EE}99.27 & \cellcolor[HTML]{BDD7EE}97.79 & \cellcolor[HTML]{5B9BD5}100.00 & \cellcolor[HTML]{BDD7EE}99.02$_{\pm0.07}$ \\
 &  & \textbf{T-Patcher} & 90.81 & 44.73 & 89.42 & 74.99$_{\pm0.23}$ & 89.34 & 14.03 & 51.69$_{\pm0.16}$ & 89.87 & 31.92 & 87.83 & 69.87$_{\pm0.31}$ & 93.16 & 92.10 & 94.26 & 93.17$_{\pm0.38}$ \\
 &  & \textbf{LEMoE} & 83.02 & 63.16 & 94.96 & 80.38$_{\pm0.30}$ & 76.42 & 37.33 & 56.88$_{\pm0.56}$ & \cellcolor[HTML]{BDD7EE}99.80 & 39.40 & 96.40 & 78.53$_{\pm0.10}$ & \cellcolor[HTML]{BDD7EE}98.75 & \cellcolor[HTML]{BDD7EE}97.31 & \cellcolor[HTML]{BDD7EE}99.65 & \cellcolor[HTML]{BDD7EE}98.57$_{\pm0.01}$ \\
 &  & \textbf{GRACE} & \cellcolor[HTML]{5B9BD5}99.88 & 46.93 & \cellcolor[HTML]{BDD7EE}99.96 & 82.26$_{\pm0.04}$ & \cellcolor[HTML]{5B9BD5}99.85 & 25.82 & 62.84$_{\pm0.12}$ & \cellcolor[HTML]{5B9BD5}100.00 & 2.63 & \cellcolor[HTML]{BDD7EE}99.87 & 67.50$_{\pm0.05}$ & \cellcolor[HTML]{BDD7EE}98.88 & 30.06 & \cellcolor[HTML]{BDD7EE}99.96 & 76.30$_{\pm0.15}$ \\
 & \multirow{-7}{*}{\textbf{DE}} & \textbf{AlphaEdit} & \cellcolor[HTML]{BDD7EE}98.46 & 57.37 & 97.00 & 84.28$_{\pm0.52}$ & 95.96 & 24.56 & 60.26$_{\pm0.34}$ & \cellcolor[HTML]{BDD7EE}99.24 & 63.28 & \cellcolor[HTML]{BDD7EE}98.64 & 87.05$_{\pm0.46}$ & \cellcolor[HTML]{BDD7EE}99.66 & 94.25 & \cellcolor[HTML]{BDD7EE}99.73 & \cellcolor[HTML]{BDD7EE}97.88$_{\pm0.23}$ \\
\cmidrule(l{2pt}r{2pt}){2-2}
 \cmidrule(l{2pt}r{2pt}){3-3}
 \cmidrule(l{2pt}r{2pt}){4-7}
 \cmidrule(l{2pt}r{2pt}){8-10}
 \cmidrule(l{2pt}r{2pt}){11-14}
 \cmidrule(l{2pt}r{2pt}){15-18}
 &  & \textbf{SERAC} & \cellcolor[HTML]{BDD7EE}99.09 & 81.63 & 86.28 & 89.00$_{\pm0.19}$ & \cellcolor[HTML]{BDD7EE}98.03 & 18.13 & 58.08$_{\pm0.16}$ & 88.94 & 43.85 & 81.30 & 71.36$_{\pm0.09}$ & \cellcolor[HTML]{BDD7EE}97.14 & \cellcolor[HTML]{BDD7EE}97.56 & \cellcolor[HTML]{BDD7EE}99.95 & \cellcolor[HTML]{BDD7EE}98.22$_{\pm0.19}$ \\
 &  & \textbf{MEND} & \cellcolor[HTML]{BDD7EE}97.21 & 68.01 & 96.23 & 87.15$_{\pm0.09}$ & \cellcolor[HTML]{BDD7EE}97.19 & 47.15 & 72.17$_{\pm0.23}$ & \cellcolor[HTML]{BDD7EE}97.50 & 41.86 & 84.18 & 74.51$_{\pm0.20}$ & \cellcolor[HTML]{BDD7EE}98.47 & 96.30 & 94.65 & 96.47$_{\pm0.28}$ \\
 &  & \textbf{IBKE} & \cellcolor[HTML]{BDD7EE}98.01 & \cellcolor[HTML]{5B9BD5}95.88 & 96.34 & \cellcolor[HTML]{5B9BD5}96.74$_{\pm0.06}$ & 96.03 & 56.32 & 76.18$_{\pm0.20}$ & 96.51 & 58.30 & 78.73 & 77.85$_{\pm0.17}$ & 96.25 & \cellcolor[HTML]{BDD7EE}98.81 & \cellcolor[HTML]{BDD7EE}98.74 & \cellcolor[HTML]{BDD7EE}97.93$_{\pm0.34}$ \\
 &  & \textbf{SERAC*} & \cellcolor[HTML]{BDD7EE}99.85 & 82.37 & 86.79 & 89.67$_{\pm0.10}$ & \cellcolor[HTML]{BDD7EE}99.70 & 48.87 & 74.28$_{\pm0.25}$ & \cellcolor[HTML]{BDD7EE}99.64 & 69.01 & 88.21 & 85.62$_{\pm0.16}$ & \cellcolor[HTML]{BDD7EE}99.37 & \cellcolor[HTML]{BDD7EE}99.02 & \cellcolor[HTML]{5B9BD5}100.00 & \cellcolor[HTML]{5B9BD5}99.46$_{\pm0.17}$ \\
 &  & \textbf{MEND*} & \cellcolor[HTML]{BDD7EE}97.04 & 66.50 & 95.62 & 86.38$_{\pm0.06}$ & \cellcolor[HTML]{BDD7EE}96.97 & 74.70 & 85.83$_{\pm0.54}$ & 96.58 & 73.49 & 91.35 & 87.14$_{\pm0.42}$ & \cellcolor[HTML]{BDD7EE}99.03 & \cellcolor[HTML]{BDD7EE}98.78 & \cellcolor[HTML]{BDD7EE}98.90 & \cellcolor[HTML]{BDD7EE}98.90$_{\pm0.32}$ \\
\multirow{-12.5}{*}{\textbf{\begin{tabular}[c]{@{}c@{}}GPT-J\\      (6B)\end{tabular}}} & \multirow{-6}{*}{\textbf{ET}} & \textbf{IBKE*} & \cellcolor[HTML]{BDD7EE}97.76 & \cellcolor[HTML]{BDD7EE}94.89 & 95.54 & \cellcolor[HTML]{BDD7EE}96.06$_{\pm0.33}$ & \cellcolor[HTML]{BDD7EE}99.05 & \cellcolor[HTML]{5B9BD5}95.11 & \cellcolor[HTML]{5B9BD5}97.08$_{\pm0.16}$ & \cellcolor[HTML]{BDD7EE}97.18 & \cellcolor[HTML]{5B9BD5}85.32 & 94.62 & \cellcolor[HTML]{5B9BD5}92.37$_{\pm0.34}$ & \cellcolor[HTML]{BDD7EE}98.81 & \cellcolor[HTML]{5B9BD5}99.40 & \cellcolor[HTML]{BDD7EE}99.14 & \cellcolor[HTML]{BDD7EE}99.12$_{\pm0.16}$ \\
\cmidrule(l{2pt}r{2pt}){1-1}
\cmidrule(l{2pt}r{2pt}){2-2}
 \cmidrule(l{2pt}r{2pt}){3-3}
 \cmidrule(l{2pt}r{2pt}){4-7}
 \cmidrule(l{2pt}r{2pt}){8-10}
 \cmidrule(l{2pt}r{2pt}){11-14}
 \cmidrule(l{2pt}r{2pt}){15-18}
 &  & \textbf{W/O} & 28.36 & 48.50 & \cellcolor[HTML]{5B9BD5}100.00 & 58.95$_{\pm0.04}$ & 32.68 & 23.65 & 28.16$_{\pm0.03}$ & 0.67 & 2.76 & \cellcolor[HTML]{5B9BD5}100.00 & 34.48$_{\pm0.04}$ & 23.53 & 34.99 & \cellcolor[HTML]{5B9BD5}100.00 & 52.84$_{\pm0.06}$ \\
 &  & \textbf{FT} & 92.95 & 69.03 & 68.36 & 76.78$_{\pm0.17}$ & \cellcolor[HTML]{BDD7EE}99.42 & 35.36 & 67.39$_{\pm0.30}$ & \cellcolor[HTML]{BDD7EE}99.95 & \cellcolor[HTML]{BDD7EE}86.25 & 40.77 & 75.66$_{\pm0.30}$ & \cellcolor[HTML]{5B9BD5}99.86 & \cellcolor[HTML]{5B9BD5}99.84 & 95.48 & \cellcolor[HTML]{BDD7EE}98.39$_{\pm0.04}$ \\
 &  & \textbf{ROME} & 92.54 & 66.13 & 95.67 & 84.78$_{\pm0.34}$ & 88.79 & 20.29 & 54.54$_{\pm0.40}$ & \cellcolor[HTML]{BDD7EE}98.79 & 39.42 & \cellcolor[HTML]{BDD7EE}98.79 & 79.00$_{\pm0.36}$ & \cellcolor[HTML]{BDD7EE}98.25 & 87.07 & \cellcolor[HTML]{BDD7EE}99.72 & 95.01$_{\pm0.10}$ \\
 &  & \textbf{T-Patcher} & 84.84 & 69.45 & 93.67 & 82.65$_{\pm0.10}$ & 91.47 & 26.78 & 59.12$_{\pm0.14}$ & 95.14 & 25.29 & 62.73 & 61.05$_{\pm0.22}$ & 88.92 & 90.40 & \cellcolor[HTML]{BDD7EE}99.04 & 92.79$_{\pm0.23}$ \\
 &  & \textbf{LEMoE} & 79.77 & 69.36 & 92.08 & 80.40$_{\pm0.10}$ & 62.74 & 35.11 & 48.93$_{\pm0.49}$ & \cellcolor[HTML]{BDD7EE}99.87 & 50.65 & 96.55 & 82.36$_{\pm0.05}$ & \cellcolor[HTML]{BDD7EE}98.91 & \cellcolor[HTML]{BDD7EE}99.15 & \cellcolor[HTML]{BDD7EE}99.77 & \cellcolor[HTML]{BDD7EE}99.27$_{\pm0.06}$ \\
 &  & \textbf{GRACE} & \cellcolor[HTML]{BDD7EE}98.67 & 68.79 & \cellcolor[HTML]{BDD7EE}99.99 & 89.15$_{\pm0.11}$ & \cellcolor[HTML]{5B9BD5}99.92 & 23.80 & 61.86$_{\pm0.08}$ & \cellcolor[HTML]{5B9BD5}99.96 & 2.77 & \cellcolor[HTML]{BDD7EE}99.99 & 67.58$_{\pm0.13}$ & \cellcolor[HTML]{BDD7EE}98.73 & 35.13 & \cellcolor[HTML]{BDD7EE}99.98 & 77.95$_{\pm0.08}$ \\
 & \multirow{-7}{*}{\textbf{DE}} & \textbf{AlphaEdit} & 93.87 & 64.92 & \cellcolor[HTML]{BDD7EE}98.95 & 85.91$_{\pm0.17}$ & 95.30 & 24.13 & 59.71$_{\pm0.59}$ & \cellcolor[HTML]{BDD7EE}99.03 & 37.31 & \cellcolor[HTML]{BDD7EE}99.82 & 78.72$_{\pm0.22}$ & 95.79 & 80.25 & \cellcolor[HTML]{BDD7EE}99.59 & 91.88$_{\pm0.22}$ \\
\cmidrule(l{2pt}r{2pt}){2-2}
 \cmidrule(l{2pt}r{2pt}){3-3}
 \cmidrule(l{2pt}r{2pt}){4-7}
 \cmidrule(l{2pt}r{2pt}){8-10}
 \cmidrule(l{2pt}r{2pt}){11-14}
 \cmidrule(l{2pt}r{2pt}){15-18}
 &  & \textbf{SERAC} & \cellcolor[HTML]{BDD7EE}99.23 & 82.87 & 86.46 & 89.52$_{\pm0.04}$ & \cellcolor[HTML]{BDD7EE}99.53 & 11.78 & 55.66$_{\pm0.33}$ & 93.54 & 26.32 & 88.11 & 69.32$_{\pm0.30}$ & \cellcolor[HTML]{BDD7EE}97.43 & 91.16 & \cellcolor[HTML]{BDD7EE}99.84 & 96.14$_{\pm0.13}$ \\
 &  & \textbf{MEND} & 94.27 & 71.36 & \cellcolor[HTML]{BDD7EE}97.89 & 87.84$_{\pm0.03}$ & 96.51 & 38.02 & 67.26$_{\pm0.28}$ & \cellcolor[HTML]{BDD7EE}97.73 & 44.75 & 84.16 & 75.55$_{\pm0.09}$ & 96.11 & \cellcolor[HTML]{BDD7EE}97.43 & 96.83 & \cellcolor[HTML]{BDD7EE}96.79$_{\pm0.20}$ \\
 &  & \textbf{IBKE} & \cellcolor[HTML]{BDD7EE}97.65 & \cellcolor[HTML]{BDD7EE}96.51 & 93.83 & \cellcolor[HTML]{BDD7EE}96.00$_{\pm0.15}$ & 93.83 & 52.90 & 73.37$_{\pm0.27}$ & \cellcolor[HTML]{BDD7EE}97.95 & 59.86 & 76.47 & 78.09$_{\pm0.15}$ & 94.79 & 96.72 & \cellcolor[HTML]{BDD7EE}98.71 & \cellcolor[HTML]{BDD7EE}96.74$_{\pm0.05}$ \\
 &  & \textbf{SERAC*} & \cellcolor[HTML]{5B9BD5}99.52 & 84.53 & 88.63 & 90.89$_{\pm0.10}$ & \cellcolor[HTML]{BDD7EE}99.69 & 45.70 & 72.69$_{\pm0.11}$ & \cellcolor[HTML]{BDD7EE}99.87 & \cellcolor[HTML]{BDD7EE}85.58 & 91.93 & \cellcolor[HTML]{BDD7EE}92.46$_{\pm0.05}$ & \cellcolor[HTML]{BDD7EE}99.61 & \cellcolor[HTML]{BDD7EE}98.00 & \cellcolor[HTML]{BDD7EE}99.94 & \cellcolor[HTML]{BDD7EE}99.18$_{\pm0.14}$ \\
 &  & \textbf{MEND*} & 94.65 & 72.09 & \cellcolor[HTML]{BDD7EE}97.14 & 87.96$_{\pm0.45}$ & \cellcolor[HTML]{BDD7EE}97.12 & 74.90 & 86.01$_{\pm0.34}$ & 96.51 & 70.48 & 91.39 & 86.12$_{\pm0.06}$ & \cellcolor[HTML]{BDD7EE}98.05 & \cellcolor[HTML]{BDD7EE}98.10 & \cellcolor[HTML]{BDD7EE}98.54 & \cellcolor[HTML]{BDD7EE}98.23$_{\pm0.03}$ \\
\multirow{-12.5}{*}{\textbf{\begin{tabular}[c]{@{}c@{}}Qwen3\\      (1.7B)\end{tabular}}} & \multirow{-6}{*}{\textbf{ET}} & \textbf{IBKE*} & \cellcolor[HTML]{BDD7EE}97.72 & \cellcolor[HTML]{5B9BD5}97.64 & 95.27 & \cellcolor[HTML]{5B9BD5}96.88$_{\pm0.26}$ & \cellcolor[HTML]{BDD7EE}97.54 & \cellcolor[HTML]{5B9BD5}95.00 & \cellcolor[HTML]{5B9BD5}96.27$_{\pm0.27}$ & \cellcolor[HTML]{BDD7EE}98.66 & \cellcolor[HTML]{5B9BD5}88.04 & 93.75 & \cellcolor[HTML]{5B9BD5}93.48$_{\pm0.22}$ & \cellcolor[HTML]{BDD7EE}99.20 & \cellcolor[HTML]{BDD7EE}99.55 & \cellcolor[HTML]{BDD7EE}99.52 & \cellcolor[HTML]{5B9BD5}99.42$_{\pm0.21}$ \\
\cmidrule(l{2pt}r{2pt}){1-1}
\cmidrule(l{2pt}r{2pt}){2-2}
 \cmidrule(l{2pt}r{2pt}){3-3}
 \cmidrule(l{2pt}r{2pt}){4-7}
 \cmidrule(l{2pt}r{2pt}){8-10}
 \cmidrule(l{2pt}r{2pt}){11-14}
 \cmidrule(l{2pt}r{2pt}){15-18}
 &  & \textbf{W/O} & 32.54 & 45.70 & \cellcolor[HTML]{5B9BD5}100.00 & 59.41$_{\pm0.06}$ & 36.26 & 25.01 & 30.64$_{\pm0.07}$ & 0.82 & 2.30 & \cellcolor[HTML]{5B9BD5}100.00 & 34.37$_{\pm0.01}$ & 27.65 & 38.03 & \cellcolor[HTML]{5B9BD5}100.00 & 55.23$_{\pm0.02}$ \\
 &  & \textbf{FT} & \cellcolor[HTML]{BDD7EE}99.83 & 68.90 & 96.39 & 88.37$_{\pm0.06}$ & \cellcolor[HTML]{BDD7EE}99.99 & 32.07 & 66.03$_{\pm0.07}$ & \cellcolor[HTML]{BDD7EE}99.85 & 32.29 & 88.89 & 73.68$_{\pm0.08}$ & \cellcolor[HTML]{BDD7EE}99.73 & 95.44 & \cellcolor[HTML]{BDD7EE}99.64 & \cellcolor[HTML]{BDD7EE}98.27$_{\pm0.19}$ \\
 &  & \textbf{ROME} & 94.36 & 64.85 & 96.44 & 85.21$_{\pm0.34}$ & \cellcolor[HTML]{BDD7EE}98.35 & 23.90 & 61.13$_{\pm0.49}$ & \cellcolor[HTML]{BDD7EE}99.50 & 46.07 & \cellcolor[HTML]{BDD7EE}99.61 & 81.73$_{\pm0.24}$ & \cellcolor[HTML]{BDD7EE}98.63 & 91.55 & \cellcolor[HTML]{BDD7EE}99.46 & \cellcolor[HTML]{BDD7EE}96.55$_{\pm0.23}$ \\
 &  & \textbf{T-Patcher} & 87.60 & 49.18 & 92.69 & 76.49$_{\pm0.31}$ & 85.79 & 25.27 & 55.53$_{\pm0.31}$ & 88.86 & 27.29 & 79.78 & 65.31$_{\pm0.23}$ & 88.69 & 86.84 & 94.74 & 90.09$_{\pm0.25}$ \\
 &  & \textbf{LEMoE} & 81.74 & 63.87 & 95.76 & 80.45$_{\pm0.27}$ & 68.79 & 34.64 & 51.71$_{\pm0.17}$ & \cellcolor[HTML]{BDD7EE}99.88 & 42.65 & \cellcolor[HTML]{BDD7EE}98.20 & 80.24$_{\pm0.05}$ & \cellcolor[HTML]{5B9BD5}99.92 & \cellcolor[HTML]{5B9BD5}98.40 & \cellcolor[HTML]{BDD7EE}99.94 & \cellcolor[HTML]{5B9BD5}99.42$_{\pm0.09}$ \\
 &  & \textbf{GRACE} & \cellcolor[HTML]{5B9BD5}99.88 & 64.05 & \cellcolor[HTML]{BDD7EE}99.99 & 87.97$_{\pm0.10}$ & \cellcolor[HTML]{5B9BD5}100.00 & 25.93 & 62.96$_{\pm0.08}$ & \cellcolor[HTML]{5B9BD5}99.90 & 2.29 & \cellcolor[HTML]{BDD7EE}99.99 & 67.39$_{\pm0.14}$ & \cellcolor[HTML]{BDD7EE}99.36 & 38.01 & \cellcolor[HTML]{BDD7EE}99.95 & 79.11$_{\pm0.04}$ \\
 & \multirow{-7}{*}{\textbf{DE}} & \textbf{AlphaEdit} & 95.74 & 64.16 & \cellcolor[HTML]{BDD7EE}99.12 & 86.34$_{\pm0.09}$ & 91.55 & 23.51 & 57.53$_{\pm0.58}$ & \cellcolor[HTML]{BDD7EE}99.52 & 44.96 & \cellcolor[HTML]{BDD7EE}99.81 & 81.43$_{\pm0.28}$ & \cellcolor[HTML]{BDD7EE}97.95 & 85.30 & \cellcolor[HTML]{BDD7EE}99.61 & 94.29$_{\pm0.24}$ \\
\cmidrule(l{2pt}r{2pt}){2-2}
 \cmidrule(l{2pt}r{2pt}){3-3}
 \cmidrule(l{2pt}r{2pt}){4-7}
 \cmidrule(l{2pt}r{2pt}){8-10}
 \cmidrule(l{2pt}r{2pt}){11-14}
 \cmidrule(l{2pt}r{2pt}){15-18}
 &  & \textbf{SERAC} & \cellcolor[HTML]{BDD7EE}98.24 & 81.79 & 83.50 & 87.84$_{\pm0.21}$ & \cellcolor[HTML]{BDD7EE}98.69 & 14.65 & 56.67$_{\pm0.16}$ & 89.76 & 41.63 & 82.51 & 71.30$_{\pm0.10}$ & 94.25 & 89.87 & \cellcolor[HTML]{BDD7EE}99.76 & 94.63$_{\pm0.20}$ \\
 &  & \textbf{MEND} & 94.45 & 71.23 & 95.44 & 87.04$_{\pm0.11}$ & 93.29 & 38.44 & 65.86$_{\pm0.17}$ & 92.98 & 45.35 & 83.54 & 73.96$_{\pm0.36}$ & 95.95 & \cellcolor[HTML]{BDD7EE}96.30 & 96.26 & 96.17$_{\pm0.14}$ \\
 &  & \textbf{IBKE} & \cellcolor[HTML]{BDD7EE}97.12 & \cellcolor[HTML]{5B9BD5}96.23 & \cellcolor[HTML]{BDD7EE}97.41 & \cellcolor[HTML]{BDD7EE}96.92$_{\pm0.19}$ & 91.87 & 47.85 & 69.86$_{\pm0.20}$ & 91.66 & 55.03 & 77.47 & 74.72$_{\pm0.25}$ & 93.51 & 94.68 & 96.75 & 94.98$_{\pm0.28}$ \\
 &  & \textbf{SERAC*} & \cellcolor[HTML]{BDD7EE}99.54 & 83.82 & 85.65 & 89.67$_{\pm0.22}$ & \cellcolor[HTML]{BDD7EE}99.72 & 46.79 & 73.26$_{\pm0.05}$ & \cellcolor[HTML]{BDD7EE}98.88 & 76.00 & 88.83 & 87.90$_{\pm0.24}$ & \cellcolor[HTML]{BDD7EE}99.08 & \cellcolor[HTML]{BDD7EE}97.47 & \cellcolor[HTML]{BDD7EE}99.67 & \cellcolor[HTML]{BDD7EE}98.74$_{\pm0.22}$ \\
 &  & \textbf{MEND*} & 95.08 & 72.18 & 95.27 & 87.51$_{\pm0.32}$ & 93.83 & 73.24 & 83.53$_{\pm0.25}$ & 94.90 & 65.16 & 89.59 & 83.22$_{\pm0.11}$ & 95.85 & \cellcolor[HTML]{BDD7EE}96.86 & \cellcolor[HTML]{BDD7EE}97.98 & \cellcolor[HTML]{BDD7EE}96.90$_{\pm0.22}$ \\
\multirow{-12.5}{*}{\textbf{\begin{tabular}[c]{@{}c@{}}Qwen3\\      (8B)\end{tabular}}} & \multirow{-6}{*}{\textbf{ET}} & \textbf{IBKE*} & \cellcolor[HTML]{BDD7EE}97.48 & \cellcolor[HTML]{BDD7EE}96.20 & \cellcolor[HTML]{BDD7EE}97.43 & \cellcolor[HTML]{5B9BD5}97.03$_{\pm0.19}$ & 96.00 & \cellcolor[HTML]{5B9BD5}88.75 & \cellcolor[HTML]{5B9BD5}92.38$_{\pm0.20}$ & 96.34 & \cellcolor[HTML]{5B9BD5}84.26 & 93.77 & \cellcolor[HTML]{5B9BD5}91.46$_{\pm0.22}$ & \cellcolor[HTML]{BDD7EE}97.64 & \cellcolor[HTML]{BDD7EE}97.16 & \cellcolor[HTML]{BDD7EE}98.36 & \cellcolor[HTML]{BDD7EE}97.72$_{\pm0.15}$\\
\bottomrule

    \end{tabularx}
    \end{threeparttable}
    \end{center}
    \label{tab_overall_edit_performance}
\end{table*}

\noindent\textbf{Benchmarks:}
To ensure a rigorous evaluation, datasets were selected according to criteria such as the complexity of edit generalization, dataset scale, and the inclusion of counterfactual knowledge.
Specifically, we employ four widely used benchmarks: ZSRE~\cite{ZSRE, MEND}, CounterFact~\cite{ROME}, MQuAKE~\cite{MQuAKE}, and UniEdit~\cite{UniEdit}.

ZSRE is a question answering (QA) dataset in which generalization samples are created by back-translation, resulting in paraphrased questions that form equivalence neighborhoods. 
CounterFact is a more challenging benchmark consisting of counterfactual statements that typically receive lower confidence scores from LLMs than their factual counterparts.
MQuAKE is designed to evaluate whether edited models can successfully handle multi-hop reasoning questions related to the edited facts. 
UniEdit is a large-scale, open-domain benchmark based on Wikidata~\cite{wikidata}, covering 25 human domains across five major sectors: Natural Sciences, Humanities, Social Sciences, Applied Sciences, and Interdisciplinary Studies. It utilizes a unified multi-hop triple-chain sampling algorithm to construct diverse and comprehensive evaluation scenarios.

For all datasets except MQuAKE, we followed the original train–test splits. 
Because MQuAKE does not include a designated training set, we sampled one-sixth of its data for use in few-shot augmentation. Statistical details for each benchmark are presented in Table~\ref{tab_statistics_benchmarks}.

\noindent\textbf{Training Settings:}
IBKE, MEND, and SERAC were trained using the UniEdit training set.
To assess the effect of small-sample augmentation, we additionally included 512 samples from each of the training sets of CounterFact, ZSRE, and MQuAKE. 
Results obtained with this augmentation are indicated with an asterisk (*).
Early stopping was applied if validation loss ceased to decrease for 50{,}000 steps, and the total number of training steps was capped at 1{,}000{,}000.
Model checkpoints were saved every 3{,}000 iterations, with the checkpoint achieving the lowest validation loss selected for final evaluation.
Each training run required approximately 3--7 days on two NVIDIA A6000 GPUs (with a single GPU used for training GPT2-XL and Qwen3-1.7B).

\noindent\textbf{Evaluation:}
Following the procedure described in~\cite{UniEdit}, we compute three evaluation metrics for model editing: reliability, generality, and locality.
For the reliability score, we measure the top-1 token-level hit rate of the post-edited LLM on the designated edit targets.
Generality is assessed by analyzing the predicted probability distribution over the target tokens and determining whether each token appears among the top five predictions (the top-5 hit rate), which reflects how effectively the post-edited LLM improves recall of the intended concepts.
For locality, we first record the original predictions of the model on the locality samples and then evaluate the top-5 hit rate for these samples after model editing.

\begin{figure*}[!t]
    \centering
    \includegraphics[width=1\textwidth]{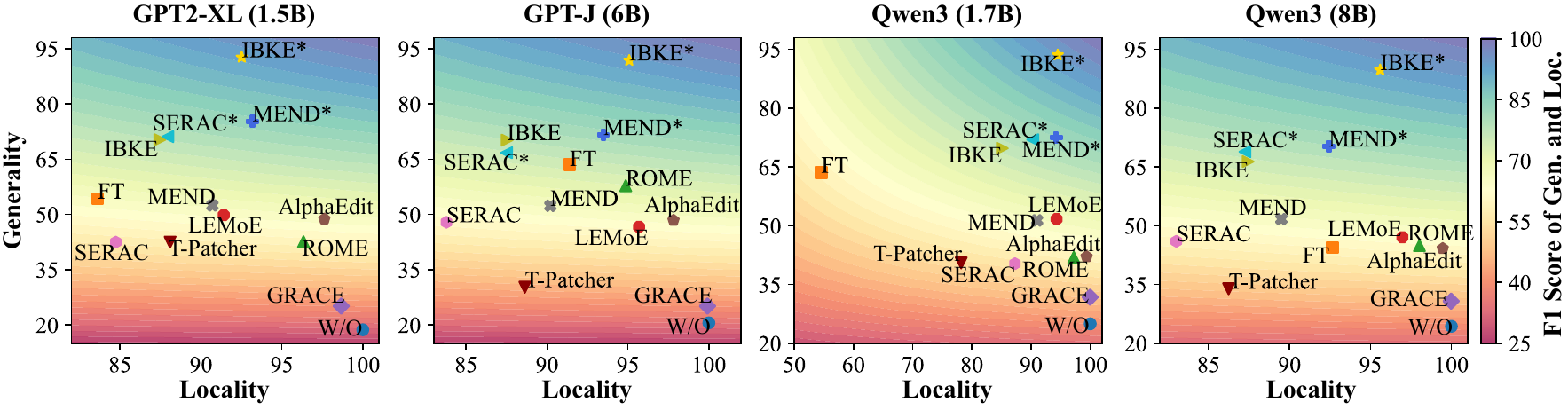}
\caption{
The trade-off between generality and locality is illustrated for various editing methods and model backbones. 
Each data point represents the average generality and locality scores achieved by a given editor, computed as the mean over the UniEdit, MQuAKE, and CounterFact benchmark datasets.
}
\label{fig_gen_loc_f1}
\end{figure*}

\subsection{Evaluation of Overall Editing Performance}
\label{sec_overall_performance}

Table~\ref{tab_overall_edit_performance} summarizes the performance of various editing methods across a range of benchmarks and LLM backbones.

Direct Editing (DE)-based methods consistently achieve high reliability and locality scores. By localizing edits to sensitive model weights and employing low-rank adaptation, these approaches enforce strong edit confidence while minimizing disruption to unrelated knowledge. On the ZSRE dataset, where generalization primarily involves paraphrasing of edit samples, DE-based methods also perform strongly. As ZSRE is relatively straightforward, most methods achieve similarly high scores across the three metrics (within 3\% of the top results). However, in counterfactual editing scenarios, such as those presented by CounterFact, these methods exhibit low generality, struggling to extend edits beyond direct instances. On more complex datasets, including UniEdit and MQuAKE, which feature diverse edit types, multi-hop reasoning, and interactions among multiple edits, DE-based approaches yield notably low generality scores. These findings indicate that DE-based methods tend to enhance prediction for directly edited instances without fostering broader integration of new knowledge within the LLM. Consequently, the model cannot reliably propagate edited knowledge to address indirect or ripple effects.

In contrast, Edit Training (ET)-based methods employ edit-specific training to establish interactions between newly introduced knowledge and the model's internal representations, leading to improved generality, as evidenced by their overall performance on UniEdit and ZSRE. However, because training is typically performed on real-world data, generality remains constrained for counterfactual datasets such as MQuAKE and CounterFact, which demand stronger intervention to override entrenched model priors. Remarkably, supplementing ET-based methods with small-sample data leads to substantial improvements in generality. 

Specifically, IBKE* utilizes holistic representations of edit requests and integrates the IB mechanism to distill essential information, thereby achieving effective generalization while minimizing interference with unrelated knowledge. By comparison, MEND relies solely on an MLP for token-level gradient decomposition and lacks a unified approach to edit semantics; this often results in excessive editing of individual samples and diminished generalization. SERAC, similarly, is constrained by the capacity of its counterfactual sub-model and its single-retrieval design, which limits its ability to support complex interactions among multiple edits.

Figure~\ref{fig_gen_loc_f1} illustrates the trade-off between generality and locality, providing an intuitive evaluation of each editor's ability to calibrate LLM responses to edited inputs while preserving performance on unrelated queries. Scores are averaged over three discriminative benchmarks (excluding ZSRE). Analogous to recall and precision in classification tasks, the calculated \textit{F1 score} (the harmonic mean of generality and locality) quantitatively reflects how precisely the editing method delineates the boundaries of knowledge modification. DE-based methods tend to cluster within or below the yellow band, indicating high locality but limited generality. ET-based methods achieve a more balanced performance, particularly after few-shot augmentation, with IBKE demonstrating the best overall trade-off.

These results underscore the advantages of introducing edit priors via targeted edit training and incorporating the IB mechanism.

\begin{figure}[!t]
    \centering
    \includegraphics[width=1\columnwidth]{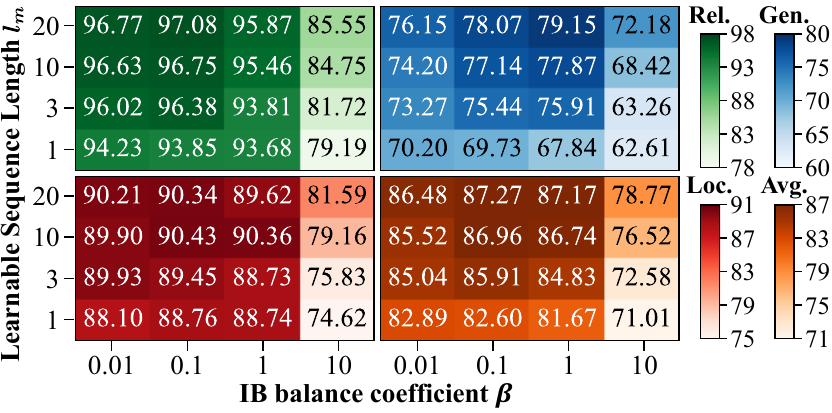}
\caption{
Hyperparameter search for the learnable sequence length $l_m$ and the IB trade-off coefficient $\beta$, using GPT2-XL as the backbone. IBKEs with different configurations are trained on UniEdit, and the results show the average performance across the four datasets.
}
\label{fig_hyper_parameter_search}
\end{figure}

\begin{figure*}[!b]
    \centering
    \includegraphics[width=1\textwidth]{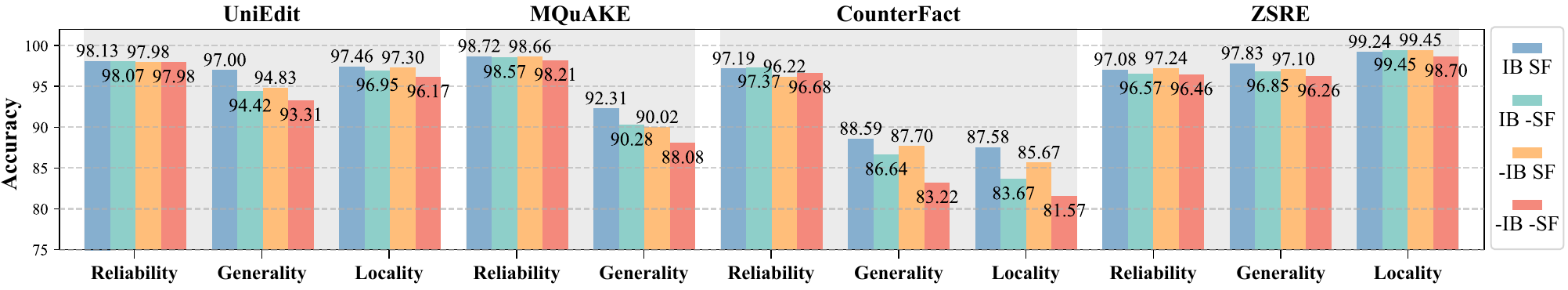}
\caption{
Ablation study of the IB mechanism and the scale factor (SF) $f_{W_{s}}(\tilde{s}_i)$, using GPT2-XL as the backbone. 
The four IBKE variants with different configurations are trained on the few-shot augmented data. Legends marked with a ``–'' sign indicate that the corresponding module has been removed.
}
\label{fig_ablation_study}
\end{figure*}

\subsection{Hyperparameter Search}
To optimize the performance of IBKE, we investigate the effects of two key hyperparameters: the learnable sequence length $l_m$ and the IB balance coefficient $\beta$, as shown in Figure~\ref{fig_hyper_parameter_search}.
Increasing $l_m$ generally leads to improved editing performance, as it directly expands the capacity of the latent representation.
The coefficient $\beta$ governs the generalization ability of the latent representation by mediating the trade-off between information compression and retention.

Unlike $l_m$, editing performance does not improve monotonically with larger $\beta$ values.
For $l_m>1$, reliability and locality typically reach their maximum at $\beta=0.1$, while generality is greatest at $\beta=1$. However, setting $\beta$ to excessively large values (e.g., $\beta=10$) diminishes all three metrics, as the latent representations become too similar and the flow of useful information is overly restricted.
Conversely, when $l_m=1$, increasing $\beta$ monotonically degrades editing performance, likely due to insufficient latent representation capacity for effective compression.

Based on these observations, our selection strategy is as follows: employ a larger $l_m$ and choose $\beta=0.1$ or $\beta=1$, depending on the desired degree of generalization.
Higher values of $\beta$ enhance generality but may reduce the model’s fidelity to the original edit (as measured by top-1 hit rate).
To balance performance and computational efficiency, we select $l_m=10$ and $\beta=0.1$ for the remaining experiments in this study.

\subsection{Ablation Study}
Figure~\ref{fig_ablation_study} presents the ablation study examining the roles of two key components in IBKE: the IB mechanism and the scale factor $f_{W_s}(\tilde{s}_i)$.

For reliability, removing either component leads to only minor decreases in performance. This outcome occurs because the input gradient signal is sufficiently informative to guide the editor in aligning the model with the edit request, even without deeper semantic processing or additional contextual information. This result is consistent with findings from MEND~\cite{MEND}, which also achieves high reliability through the use of multilayer perceptrons (MLPs) to independently adjust token-level gradient decompositions.
The IB mechanism improves IBKE’s capacity to interpret edit semantics and extract key information by compressing the edit signal. When omitted, redundant information remains within the latent representation, impairing generalization and making it more challenging for the scale factor to focus on the most informative token positions.

The scale factor modulates the strength of token-level gradient updates based on the extracted semantic features of the edit. By highlighting important token positions and suppressing less pertinent ones, it promotes more targeted adjustments. Its absence causes all token updates to be applied with uniform strength, resulting in diminished generality and locality.

In summary, the synergy between the IB mechanism and the scale factor enables the editor to isolate crucial edit information while filtering out unproductive updates, thereby enhancing both generality and locality.

\begin{figure*}[!t]
    \centering
    \subfloat[Overall performance across the 25 domains.]{
        \includegraphics[width=1\textwidth]{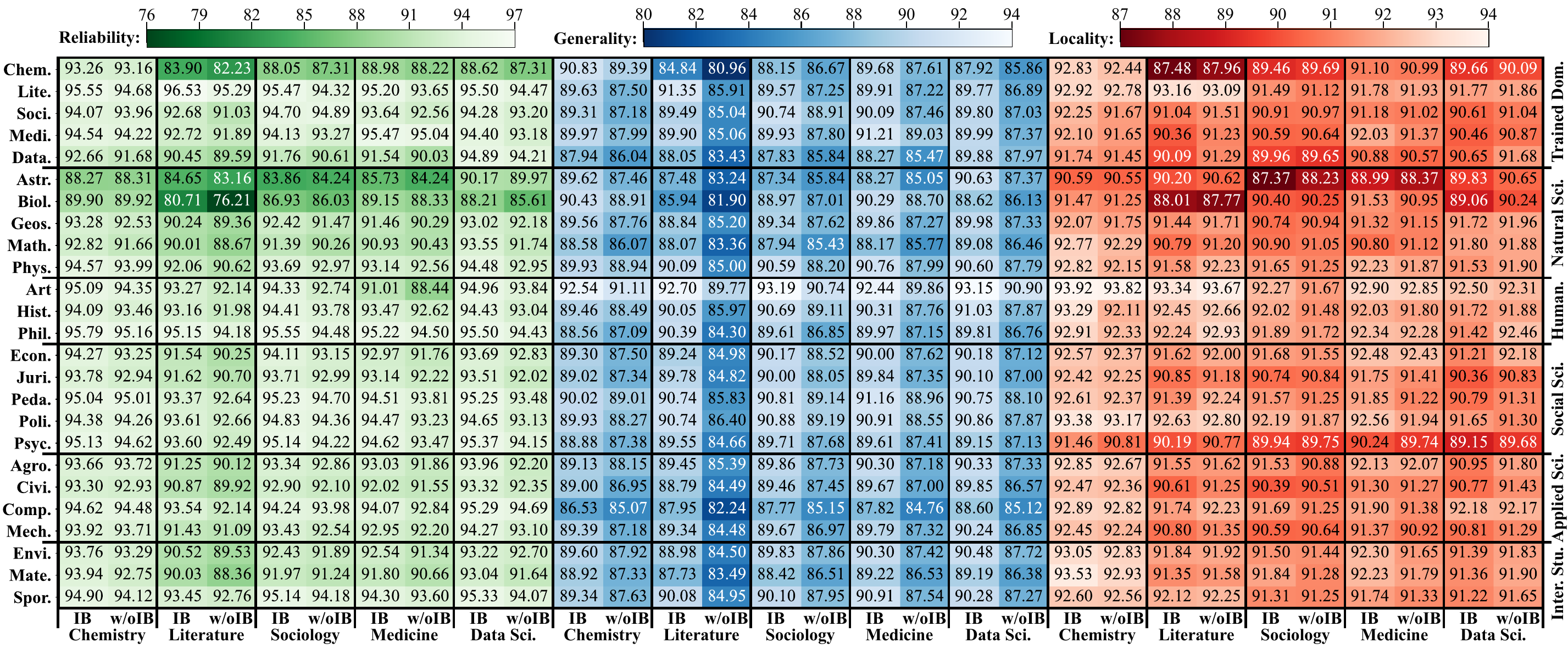}
        \label{fig_IB_ablation_across_dom}
    }
    \\
    \subfloat[Average generality across different criteria.]{
        \includegraphics[width=1\textwidth]{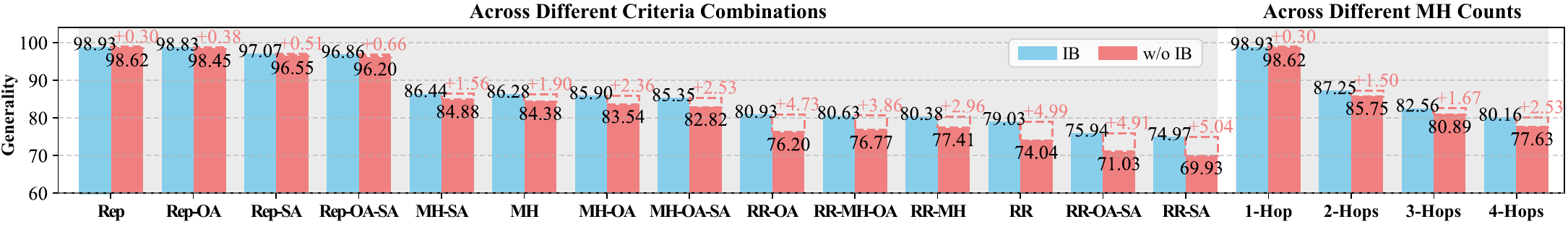}  
        \label{fig_IB_ablation_across_criteria}
    }
    \caption{
    Performance of IBKE with and without the IB mechanism, trained on five domains from different sectors in UniEdit, using Qwen3-1.7B as the backbone. 
    (a) shows the overall performance of each training instance across the 25 domains in UniEdit, where the vertical axis represents the 25 test domains, and the horizontal axis represents the training domains.
    (b) shows the average generality over the five training instances across different combinations of criteria and hop counts in UniEdit, where the abbreviations denote Rephrase (Rep), Object Alias (OA), Subject Alias (SA), Multi-Hop (MH), and Relation Reverse (RR).
    }
    \label{fig_IB_ablation_across_dom_criteria}
\end{figure*}

\subsection{Granular Performance Evaluation of the IB Mechanism}
To further investigate the performance improvement provided by the IB mechanism in different scenarios, we conducted a more granular analysis across multiple domains and evaluation criteria. As shown in Figure~\ref{fig_IB_ablation_across_dom_criteria}, each editor instance was trained on a knowledge domain containing fewer than 9K samples on average, resulting in overall performance slightly lower than that of the main experiment. 

Subfigure~\ref{fig_IB_ablation_across_dom} illustrates the enhancement effect of the IB mechanism on the editor across various knowledge domains.
For reliability and generality, editors with IB applied (on the left) generally exhibit higher background brightness, corresponding to higher scores. 
Notably, in more challenging domain adaptation scenarios, the introduction of IB tends to yield a greater enhancement effect. 
For example, the reliability difference for editors trained in the Literature domain, when evaluated on the Biology domain, reaches 4.5, while the generality difference exceeds 3.5 across nearly all domains.
This suggests that introducing IB facilitates the editor's adaptation to new edit domains. Additionally, while the effect on locality is less pronounced, editors with IB still tend to outperform in most cases.

\begin{figure}[!t]
    \centering
    \includegraphics[width=1\columnwidth]{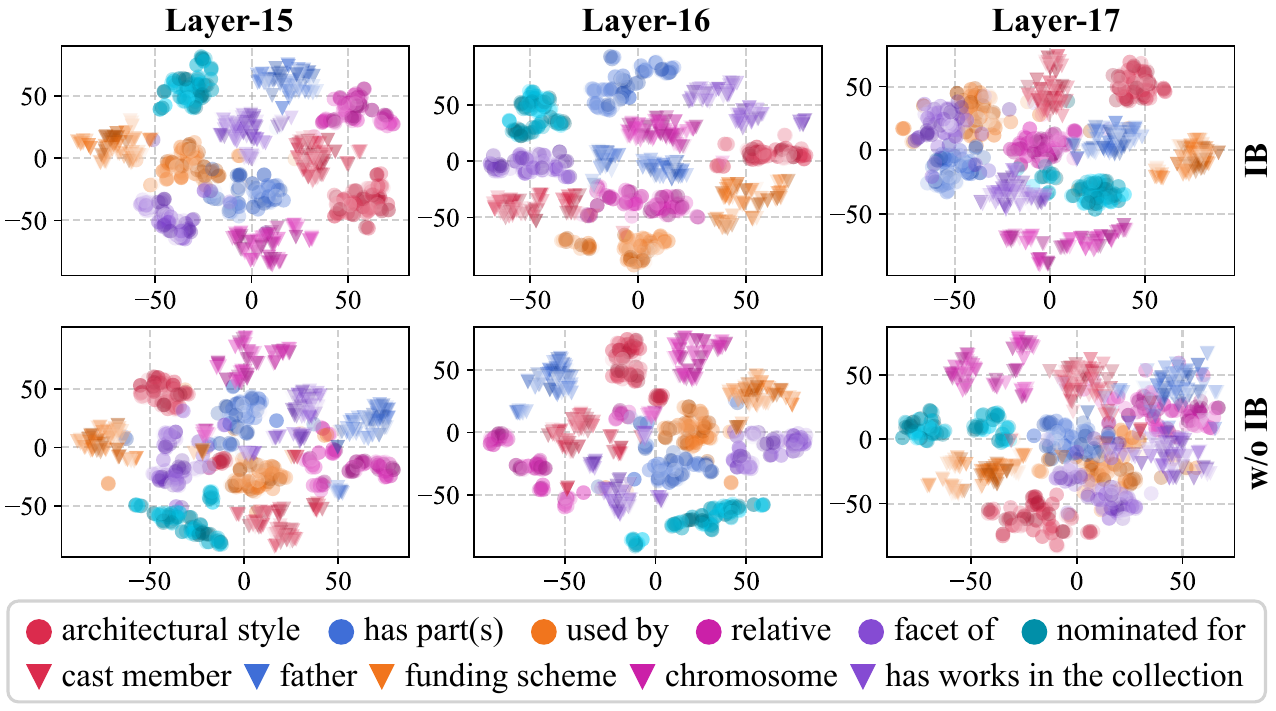}
\caption{
Latent representation visualization of IBKE with and without the IB mechanism using t-SNE, with GPT2-XL as the backbone. 
Each point represents an edit sample, and each class corresponds to samples sharing the same relation. Brightness indicates the prediction confidence of the edited model given the edited samples.
}
\label{fig_IBKE_TSNE}
\end{figure}

Subfigure~\ref{fig_IB_ablation_across_criteria} presents the enhancement effect of IB on the editor from the perspective of generality evaluation criteria. 
A clear stair-step pattern is observed, with scores dropping as each additional criterion—such as Multi-Hop and Relation Reverse—is included.
Similar to Subfigure~\ref{fig_IB_ablation_across_dom}, IB provides a stronger enhancement of the editor's generality under more difficult criteria, and a similar effect is observed as the number of hops increases.

\begin{figure*}[!t]
    \centering
    \subfloat[Scale factor values across layers for each edit token.]{
        \includegraphics[width=1\textwidth]{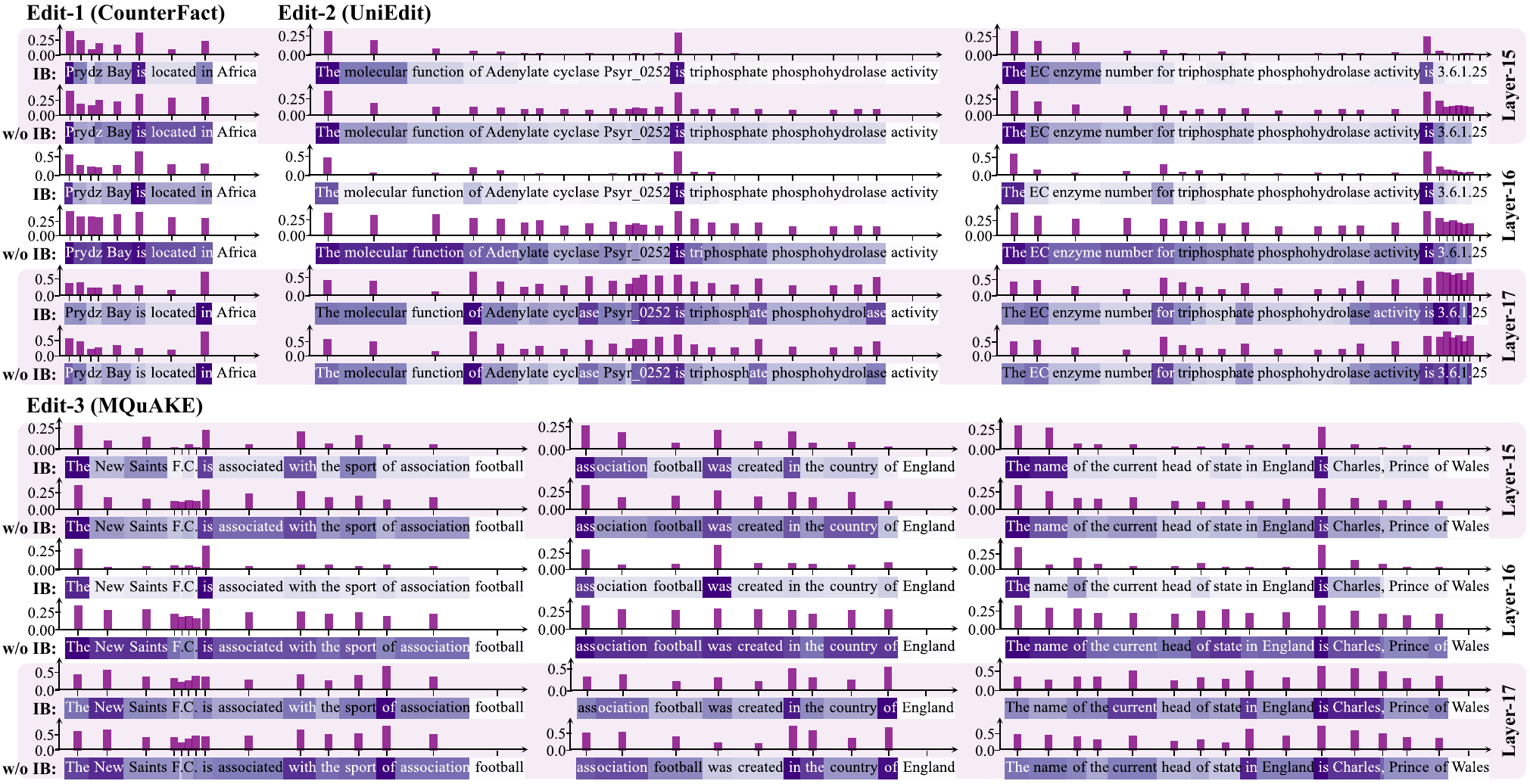}
    \label{fig_edit_token_significance_visualization_bars}
    }
    \\
    \subfloat[Changes in edit metrics as scale factors are pruned.]{
        \includegraphics[width=1\textwidth]{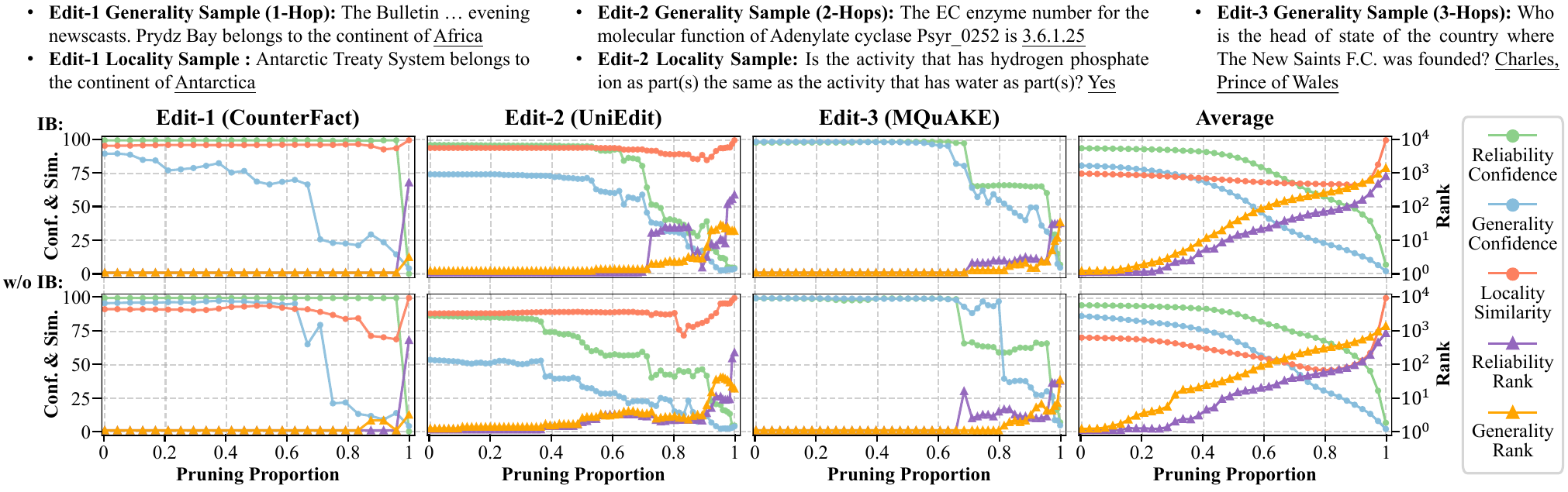} 
    \label{fig_edit_token_significance_visualization_lines}
    }
    \caption{
Visualization of edit token importance for IBKE with and without the IB mechanism, using GPT-2-XL as the backbone. The three edit examples are drawn from CounterFact, UniEdit, and MQuAKE, respectively.
(a) shows the strength of scale factors $f_{W_{s}}(\tilde{s}_i)$ across layers for each edit token (values are activated by the sigmoid function), where darker token backgrounds intuitively correspond to higher values.  
(b) displays changes in edit metrics as scale factors are iteratively pruned to zero, from the lowest to the highest values. 
Both the average prediction confidence and the vocabulary rank of the target tokens are reported for reliability and generality.
Locality similarity is computed based on the JS divergence between the edited and original models’ prediction distributions on the locality samples.
``Average'' reports the averaged results over 3,000 successfully edited instances randomly drawn from the three datasets.
}
    \label{fig_edit_token_significance_visualization}
\end{figure*}

\subsection{Feature Visualization and Instance Analysis}
To illustrate the semantic representation learning capabilities of our editor architecture, we visualize the distribution of its latent representations—extracted by the learnable fixed-length sequence—using dimensionality reduction via t-SNE~\cite{t-SNE}, as shown in Figure~\ref{fig_IBKE_TSNE}.
We observe that, regardless of whether the IB mechanism is employed, edit samples sharing the same relation tend to cluster together, indicating that both editor variants effectively capture the distribution of edit semantics through the learned representations.
In layers 15 and 16, these clusters are well separated, whereas in layer 17, some clusters exhibit less distinct boundaries, particularly for the relations ``has part(s)'', ``used by'', and ``facet of''. 
We hypothesize that this results from differences in the semantic level encoded by the gradient decomposition of GPT2-XL within this layer.
Moreover, incorporating the IB mechanism produces clearer boundaries between clusters and fewer outliers, suggesting improved learning of distinct and well-separated latent representations.

Figure~\ref{fig_edit_token_significance_visualization} further visualizes the scale factors $f_{W_s}(\tilde{s}_i)$ to assess whether IBKE attends to key positions in the input edit signals, and to examine how the addition of the IB mechanism affects this process.
Subfigure~\ref{fig_edit_token_significance_visualization_bars} shows the distribution of scale factors for edit samples from three different datasets, processed by editor instances with and without the IB mechanism.
At the token level, we find that the first token, subject, predicate, and prepositions within an edit sentence receive varying attention across layers.
Scale factors in layers 15 and 16 exhibit greater sparsity, while those in layer 17 are more densely distributed. This pattern indicates that these key positions convey most of the semantics necessary for effective editing.
Comparing editor instances, we find that scale factors in models trained with the IB mechanism are noticeably sparser, suggesting that IB-driven compact representations enhance the sparsity and selectivity of gradient decompositions.

Subfigure~\ref{fig_edit_token_significance_visualization_lines} depicts how editing performance changes as scale factors are systematically pruned from the smallest to largest values, providing an intuitive demonstration of the advantages of increased sparsity.
Editors utilizing IB show more moderated confidence in both reliability and generality, while those lacking IB may over-edit or affect unrelated samples, as reflected by reduced locality similarity.
Additionally, as shown in the ``Average'' column, editors with IB exhibit slower increases in prediction rank for generality samples during early pruning, whereas editors without IB experience a pronounced drop in locality similarity during later stages.
Overall, these results suggest that the IB mechanism mitigates interference from redundant, entangled token positions during editing, contributing to greater robustness and superior overall performance.

\section{Discussion}
Current editors for LLMs are often effective within narrowly defined or constrained settings, but they frequently struggle to generalize in open, real-world scenarios~\cite{ROME, MEMIT, T-Patcher, LEMOE, GRACE}. 
In this work, we introduce a theoretically principled framework for model editing that explicitly manages the balance between generality and locality.
By framing model editing as an information-constrained optimization problem, the approach distinguishes essential edit information from redundant signals, enabling effective generalization of edits while reducing interference with unrelated knowledge.

Building on this foundation, we develop the Information Bottleneck Knowledge Editor (IBKE), which implements the information bottleneck (IB) principle through gradient-based latent encoding and adaptive token-level scaling. 
Empirical results drawn from four benchmarks, four backbone architectures, and nine baseline methods—comprising 208 evaluation experiments in total—demonstrate that IBKE achieves leading performance in balancing generality and locality.

Extensive hyperparameter exploration reveals that a moderate balance coefficient ($\beta \approx 0.1$) strikes an optimal compromise between reliability and generality, providing new insight into how the IB principle guides model editing. 
Ablation studies demonstrate that both the IB mechanism and the token-level scale factor are critical; removal of either component results in redundant or misaligned updates. This underscores that information compression and adaptive scaling are key for IBKE to focus on semantically relevant changes and to precisely execute edits.

Granular evaluations across knowledge domains and evaluation criteria show that editors incorporating the IB mechanism demonstrate enhanced cross-domain generalization and greater robustness under challenging conditions, indicating that IB supports the learning of transferable editing patterns.
Visualization analyses further reveal that IB increases the separability of latent semantic clusters and promotes sparsity in token-level updates, confirming that the editor effectively targets and utilizes the most informative aspects of the edit. These findings support our central hypothesis: constraining information flow enables more robust, balanced, and interpretable model updating.

Overall, our study demonstrates that model editing within the IB framework offers a promising approach for achieving reliable, generalizable, and interpretable updates in large language models, with IBKE representing a significant advance in this direction. One limitation of our current implementation is the relatively high-rank nature of weight updates. Future work may seek to integrate the IB principle with low-rank adaptation strategies (e.g., rank-1 updates) to further reduce potential interference with original model parameters. In addition, extending IBKE with retrieval mechanisms for lifelong editing offers another exciting avenue to broaden its utility and impact.

\section{Methods}

In this section, we first describe the model editing scenario and provide background on the IB principle. We then introduce the IB–enhanced model editing framework and its information flow, followed by a detailed description of the  IBKE, which is constructed based on this framework.

\subsection{Preliminaries}

\subsubsection{Model Editing}

We represent an LLM as $f_{\theta} \in \mathcal{F}$, where $f_{\theta}: Q \to O$ maps an input query $q \in Q$ to an output $o = f_{\theta}(q)$.
An edit request $e = (q_e, o_e)$ instructs the LLM to produce the target output $o_e$ when given the prompt $q_e$, particularly when $f_{\theta}(q_e) \neq o_e$.
Given a set of edit requests $\mathcal{E} = \{e_i\}_{i=1}^n$, model editing seeks to construct an editor $\phi: \mathcal{F} \times \mathcal{E} \to \mathcal{F}$ that generates a post-edit LLM, denoted $f_{\hat{\theta}_{\mathcal{E}}} = \phi(f_{\theta}, \mathcal{E})$.

An effective editor should ensure that the edited LLM satisfies the following three criteria~\cite{ZJUEditSurvey2023}:

\textbf{Reliability} assesses the accuracy of the responses generated by $f_{\hat{\theta}_{\mathcal{E}}}$ for the edited samples:
\begin{gather}
\mathbb{E}_{(q_e, o_e) \in \mathcal{E}} \left[ \mathbb{I}\left(f_{\hat{\theta}_{\mathcal{E}}}(q_e) = o_e \right) \right]
\end{gather}
where $\mathbb{I}(\cdot)$ denotes the indicator function.

\textbf{Generality} measures whether $f_{\hat{\theta}_{\mathcal{E}}}$ appropriately adapts its responses to queries related to the edited samples:
\begin{gather}
\mathbb{E}_{(q_g, o_g) \sim \mathcal{G}(\mathcal{E})} \left[ \mathbb{I}\left(f_{\hat{\theta}_{\mathcal{E}}}(q_g) = o_g \right) \right]
\end{gather}
where $\mathcal{G}(\mathcal{E})$ denotes the relevant neighborhood (the ``ripple effect'') of the edit set $\mathcal{E}$.

\textbf{Locality} evaluates whether $f_{\hat{\theta}_{\mathcal{E}}}$ maintains responses consistent with the original model for queries unrelated to the edited set:
\begin{gather}
\mathbb{E}_{(q_l, o_l) \sim \mathcal{L}(\mathcal{E})} \left[ \mathbb{I}\left(f_{\hat{\theta}_{\mathcal{E}}}(q_l) = f_{\theta}(q_l) \right) \right]
\end{gather}
where $\mathcal{L}(\mathcal{E})$ refers to the set of queries unrelated to $\mathcal{E}$, specifically excluding $\mathcal{E} \cup \mathcal{G}(\mathcal{E})$.

\subsubsection{Information Bottleneck (IB)}

Consider a supervised learning task with random variables $(X, Y)$, where $X$ denotes the model input and $Y$ is the corresponding true label.
The IB principle~\cite{DBLP:journals/corr/physics-0004057, DBLP:conf/iclr/AlemiFD017} represents the model's forward propagation process $\phi$ as a Markov chain:
\begin{gather}
Y \rightarrow \underbrace{X \rightarrow Z \rightarrow \hat{Y}}_{\phi}
\end{gather}
Here, $Z$ is an intermediate representation of $X$ produced by the model, and $\hat{Y}$ is the predicted label.
This structure implies that $Y$ and $Z$ are conditionally independent given $X$, indicating that all information relevant to $Y$ must pass through $X$.

The IB objective seeks to learn a representation $Z$ that is both compressed and sufficient, by minimizing the mutual information $I(Z; X)$ while maximizing $I(Z; Y)$:
\begin{gather}
\max I(Z; Y) - \beta I(Z; X)
\end{gather}
where $\beta > 0$ is a trade-off parameter that determines the level of compression.
By balancing these two terms, the IB principle encourages the model to extract essential information from $X$ for predicting $Y$, while reducing redundant content.

\subsection{Knowledge Editing: an IB Perspective}

In this subsection, we present a unified formulation of model editing as a Markov chain, which provides the foundation for the IB approach and its associated optimization objectives.
Unlike conventional supervised learning tasks, the forward pass here consists of two distinct stages: first, intervention by the model editor in the LLM, and second, generation of responses by the edited LLM to subsequent inputs.
We describe the information flow among relevant variables as follows:
\begin{equation}
\begin{aligned}
    \begin{tikzpicture}[auto]
        \node (E) {$E$};
        \node (Z) [right of=E, node distance=1.6cm] {$Z$};
        \node (Ql) [above of=E, node distance=1cm] {$Q_l$};
        \node (Ol) [left of=Ql, node distance=1.6cm] {$O_l$};
        \node (hatOl) [right of=Ql, node distance=1.6cm] {$\hat{O}_l$};
        \node (Qg) [below of=E, node distance=1cm] {$Q_g$};
        \node (Og) [left of=Qg, node distance=1.6cm] {$O_g$};
        \node (hatOg) [right of=Qg, node distance=1.6cm] {$\hat{O}_g$};
        \draw[->] (E) -- (Z) node[midway, below] {$\phi_1$};
        \draw[->] (E) -- (Qg);
        \draw[->] (E) -- (Ql);
        \draw[->] (Ql) -- (Ol);
        \draw[->] (Ql) -- (hatOl) node[midway, below] {$f_{\theta}$};
        \draw[->] (Qg) -- (hatOg) node[midway, below] {$f_{\theta}$};
        \draw[->] (Z) -- (hatOg) node[midway, right] {$\phi_2$};
        \draw[->] (Z) -- (hatOl) node[midway, right] {$\phi_2$};
        \draw[->] (Qg) -- (Og);
        \draw[->] (E) -- (Og);
    \end{tikzpicture}
\end{aligned}
\label{eq_model_editing_graph}
\end{equation}

Within this framework, each edit induces specific distributions for generality and locality, meaning that both generality samples $(Q_g, O_g)$ and locality samples $(Q_l, O_l)$ are conditioned on the edit request $E$.
We explicitly partition the editor $\phi$ into two stages: $\phi_1$, which derives the latent representation $Z$ from the edit request $E$, and $\phi_2$, which applies $Z$ to intervene in the LLM, thereby altering its predicted response $\hat{O}$ to the input $Q$.
Based on these dependencies, we propose three constraints for the model editing process to achieve effective editing while minimizing the transmission of redundant information.

\noindent\textbf{Information Transfer Minimization (ITM):}  
The editor should compress the information contained in the representation $Z$ as much as possible. Formally, this is expressed as
\begin{align}
\min I(E; Z).
\end{align}
Expanding the mutual information yields:
\begin{align}
\sum_{e,z} p(e, z) \log \frac{p(z|e)}{p(z)}.
\end{align}
Since $p(z)$ is generally intractable, by introducing a reference distribution $r(z)$ and leveraging the non-negativity of the KL divergence, we obtain the following variational upper bound:
\begin{align}
\sum_{e,z} p(e,z) \log \frac{p(z|e)}{p(z)} 
&\leq \sum_{e,z} p(e,z) \log \frac{p(z|e)}{r(z)}.
\label{eq_upper_bound_IB}
\end{align}
In the context of Eq.~\ref{eq_model_editing_graph}, the distribution $p(z|e)$ is parameterized as $p_{\phi_1}(z|e)$.  
Thus, the objective becomes optimizing the upper bound with respect to $\phi_1$:
\begin{align}
&\min_{\phi_1} \; \sum_{e, z} p(e) p_{\phi_1}(z|e) \log \frac{p_{\phi_1}(z|e)}{r(z)}\\
\implies \;\; &\min_{\phi_1} \; \sum_{e} p(e) \, \mathrm{KL}\left[ p_{\phi_1}(z|e) \, \| \, r(z)\right]
\label{eq_IB_obj_minimization}
\end{align}

\medskip

\noindent\textbf{Sufficiency for Generality (SG):}  
The representation $Z$ and the generality input $Q_g$ should contain as much information as possible about the generality output $O_g$. This can be written and derived as follows:
\begin{align}
&\max I(Z, Q_g; O_g) \\
\implies \; & \max \sum_{z, q_g, o_g} p(z, q_g, o_g) \log \frac{p(o_g | z, q_g)}{p(o_g)} \\
\implies \; & \max \sum_{z, q_g, o_g} p(z, q_g, o_g) \log p(o_g | z, q_g) + H(O_g) \\
\propto \; & \max \sum_{z, q_g, o_g} p(z, q_g, o_g) \log p(o_g | z, q_g)
\end{align}
Since the entropy $H(O_g)$ is constant, it is dropped in the optimization.  
By replacing $p(o_g | z, q_g)$ with an approximating distribution $\xi(o_g | z, q_g)$ and using the non-negativity of the KL divergence, we obtain the following variational lower bound:  
\begin{equation}
\begin{aligned}
\sum_{z, q_g, o_g} p(z, q_g, o_g) \log p(o_g | z, q_g) \geq \\
\sum_{z, q_g, o_g} p(z, q_g, o_g) \log \xi(o_g | z, q_g)
\end{aligned}
\end{equation}
In the context of Eq.~\ref{eq_model_editing_graph}, $\xi(o_g | z, q_g)$ is modeled by the decoder formed as the composition of $\phi_2$ and $f_{\theta}$.  
Thus, the objective becomes optimizing the lower bound with respect to $\phi$, as derived below:
\begin{align}
&\max_{\phi} \sum_{z, q_g, o_g} p(z, q_g, o_g) \log \xi(o_g | z, q_g) \\
\implies \; & \max_{\phi} \sum_{z, e, q_g, o_g} p(z, e, q_g, o_g) \log \xi(o_g | z, q_g) \\
\implies \; & \max_{\phi} \sum_{z, e, q_g, o_g} p(e) \, p(q_g, o_g | e) \, p_{\phi_1}(z | e) \log \xi(o_g | z, q_g)
\label{eq_IB_obj_sufficient}
\end{align}
The derivation of Eq.~\ref{eq_IB_obj_sufficient} follows from the conditional independence of $q_g, o_g$ and $z$ given $e$, as shown in Eq.~\ref{eq_model_editing_graph}.

\medskip

\noindent\textbf{Independence for Locality (IL):}  
Given the locality input $Q_l$, the representation $Z$ should contain as little information as possible about the locality output $\hat{O}_l$ of the LLM. This is expressed as follows:
\begin{align}
&\min I(Z; \hat{O}_l \mid Q_l) \\
\implies \; & \min \sum_{z, \hat{o}_l, q_l} p(z, \hat{o}_l, q_l) \log \frac{p(\hat{o}_l \mid z, q_l)}{p(\hat{o}_l \mid q_l)}
\end{align}
In the context of Eq.~\ref{eq_model_editing_graph}, the distribution $p(\hat{o}_l \mid z, q_l)$ is parameterized as $\xi(\hat{o}_l \mid z, q_l)$.  

For the conditional distribution $p(\hat{o}_l \mid q_l)$, we replace it with the original model $f_{\theta}$, which is unaffected by $Z$, and similar to Eq.~\ref{eq_upper_bound_IB}, we obtain a variational upper bound.
The objective then becomes:
\begin{align}
&\min_\phi \sum_{z, \hat{o}_l, q_l} p(z, \hat{o}_l, q_l) \log \frac{\xi(\hat{o}_l \mid z, q_l)}{f_{\theta}(\hat{o}_l \mid q_l)} \\
\implies \; & \min_\phi \sum_{e, z, \hat{o}_l, q_l} p(z, e) \, p(\hat{o}_l \mid q_l, z, e) \, p(q_l \mid z, e)
\cdot \notag \\
& \qquad \log \frac{\xi(\hat{o}_l \mid z, q_l)}{f_{\theta}(\hat{o}_l \mid q_l)} \\
\implies \; & \min_\phi \sum_{e, z, \hat{o}_l, q_l} p(e) \, p(q_l \mid e) \, \xi(\hat{o}_l \mid q_l, z) \, p_{\phi_1}(z \mid e)
\cdot \notag \\
& \qquad \log \frac{\xi(\hat{o}_l \mid z, q_l)}{f_{\theta}(\hat{o}_l \mid q_l)}
\label{eq_IB_obj_independence}
\end{align}
In Eq.~\ref{eq_IB_obj_independence}, the terms $p(q_l \mid e)$ and $\xi(\hat{o}_l \mid q_l, z)$ arise from conditional independence assumptions, which allow certain variables to be omitted.

\begin{figure*}[!t]
    \centering
    \includegraphics[width=0.7\textwidth]{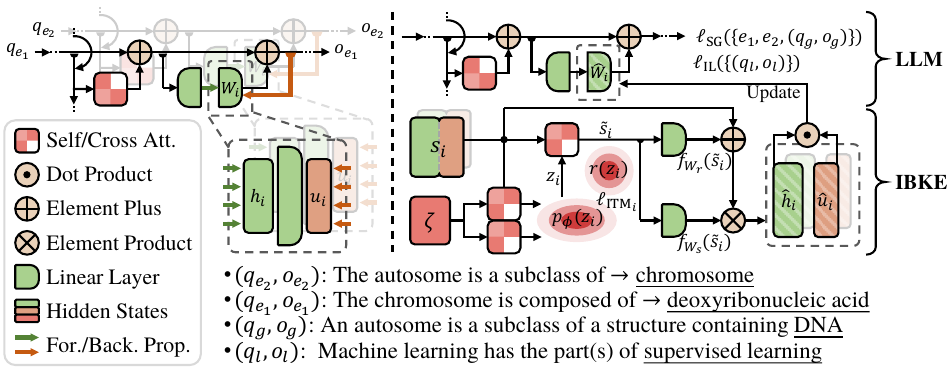}
\caption{
Overall structure of IBKE. A batch of two-hop edit samples is shown as an example. 
The left panel illustrates how edit signals are obtained, while the right panel depicts the pipeline by which IBKE utilizes these signals to update model weights.
}
\label{fig_IBKE_structure}
\end{figure*}

\subsection{Information Bottleneck Knowledge Editor (IBKE)}

In this section, we introduce IBKE, an editor built on the IB framework described above.
Inspired by previous studies on LLM knowledge localization and editing~\cite{DBLP:conf/acl/DaiDHSCW22, ROME, MEMIT, AlphaEdit}, IBKE corrects the responses of an LLM by adapting the weights of the output linear layers within the early-layer feed-forward networks (FFNs).
To achieve this, a set of hypernetworks is trained; each hypernetwork receives as input a signal that encodes the prior of the edit request and is used to generate weight offsets.
The full computational workflow of IBKE is illustrated in Figure~\ref{fig_IBKE_structure}.

\subsubsection{Obtaining Latent Representation (First Stage)}

The first stage of IBKE, denoted as $\phi_1$, computes the first-order gradient decompositions of the edit weights with respect to the edit request~\cite{MEND, DAFNet, MALMEN}, which are then used as inputs to the associated hypernetworks.
While simpler alternatives are available, such as using embeddings or hidden representations of the edit request~\cite{KnowledgeEditor, VisEdit}, the gradient decomposition conveys not only the semantics of the edit but also the direction of steepest weight adaptation.

Specifically, within the LLM $f_\theta$, we consider a set of linear layer matrices $\{W_i \in \mathbb{R}^{d_{\text{in}} \times d_{\text{out}}}\}_{i=1}^n$, each paired with a hypernetwork that processes its corresponding gradient signal.
Here, $d_{\text{in}}$ and $d_{\text{out}}$ are the input and output dimensions of each linear layer, respectively.
Given an edit request $e = (q_e, o_e)$, the gradient for each matrix can be decomposed as:
\begin{gather}
\nabla W_i = \frac{\partial \ell_s(q_e,o_e,f_\theta)}{\partial W_i} = h_i^\top u_i 
= h_i^\top \frac{\partial \ell_s(q_e,o_e,f_\theta)}{\partial (h_i W_i)}
\end{gather}
where $\ell_s$ denotes the cross-entropy loss,
$h_i \in \mathbb{R}^{l_e \times d_{\text{in}}}$ is the input to the linear layer,
and $u_i \in \mathbb{R}^{l_e \times d_{\text{out}}}$ is the gradient of the linear mapping $h_i W_i$.
The sequence length of the edit request is denoted as $l_e$.
Then, the concatenated vector $s_i = h_i \oplus u_i \in \mathbb{R}^{l_e \times (d_{\text{in}} + d_{\text{out}})}$ serves as the input to the hypernetwork.
Below, we describe the computation performed within each hypernetwork.

Given input $s_i$, IBKE models the latent distribution $p(z_i|s_i)$ and aligns it to a prior distribution $r(z_i)$ using the information transfer minimization (ITM) constraint.
To enable differentiable optimization, we employ the reparameterization trick~\cite{DBLP:journals/corr/KingmaW13, DBLP:conf/iclr/AlemiFD017}.
Specifically, $p(z_i \mid s_i)$ is parameterized as a Gaussian distribution, with two cross-attention (CA) modules mapping $s_i$ to the mean and log-variance of the distribution:
\begin{gather}
\mu_{z_i} = f_{\text{CA}_\mu}(\zeta, s_i) \in \mathbb{R}^{l_m \times d_m}, \\
v_{z_i} = \sqrt{\exp f_{\text{CA}_v}(\zeta, s_i)} \in \mathbb{R}^{l_m \times d_m}
\end{gather}  
where $\zeta \in \mathbb{R}^{l_{m} \times d_{m}}$ is a learnable sequence of fixed length $l_m$, and $d_m$ denotes the module dimension in IBKE. 

The cross-attention modules are defined as:
\begin{gather}
    f_{\text{CA}}(\zeta, s_i) = \delta\left(\zeta W_q (s_i W_k)^\top \right) s_i W_v
\end{gather}
where $W_q \in \mathbb{R}^{d_m \times d_m}$, $W_k \in \mathbb{R}^{(d_{\text{in}} + d_{\text{out}}) \times d_m}$, and $W_v \in \mathbb{R}^{(d_{\text{in}}+d_{\text{out}}) \times d_m}$ are projection matrices; biases are omitted for brevity. Here, $\delta$ denotes the softmax function.

The prior $r(z_i)$ is chosen as the standard normal distribution $\mathcal{N}(\mathbf{0}, \mathbf{I})$. 
Following Eq.~\ref{eq_IB_obj_minimization}, the ITM constraint for a request $e$ leads to the following loss, specified by the KL divergence between the two Gaussian distributions: 
\begin{equation}
\begin{aligned}
\ell_{\text{ITM}_i}(e) = \frac{1}{2}\left\{ -\mathbf{1}^\top \log \bar{v}_{z_i}^2 + \|\bar{v}_{z_i}^2\|_1 + \|\bar{\mu}_{z_i}^2\|_1 - l_m d_m \right\}
\end{aligned}
\end{equation}
where $\bar{v}_{z_i}$ and $\bar{\mu}_{z_i}$ are the flattened versions of $v_{z_i}$ and $\mu_{z_i}$, respectively, and $\mathbf{1}$ is an all-ones vector.

During training, the latent representation $z_i \in \mathbb{R}^{l_m \times d_m}$ is sampled as:
\begin{gather}
z_i = \mu_{z_i} + v_{z_i} \odot \epsilon, \;\; \epsilon \sim \mathcal{N}(\mathbf{0}, \mathbf{I})
\end{gather}
where $\epsilon \in \mathbb{R}^{l_m \times d_m}$ is noise sampled from the standard normal distribution, and $\odot$ denotes element-wise multiplication.
During inference, we set $\epsilon = 0$ to ensure deterministic outputs.

\subsubsection{Model Adaptation (Second Stage)}
In the second stage of IBKE, denoted as $\phi_2$, we adapt the LLM $f_\theta$ using the gradient decomposition augmented by the latent representation $z_i$.
Specifically, $s_i$ is reparameterized conditional on $z_i$ via a cross-attention (CA) module:
\begin{gather}
\tilde{s}_{i} = f_{\text{CA}_s}(s_i, z_i) \in \mathbb{R}^{l_e\times d_m}
\end{gather}
Next, two separate linear layers are applied to map $\tilde{s}_{i}$ into: (1) a residual vector $f_{W_r}(\tilde{s}_{i}) \in \mathbb{R}^{l_e\times (d_{\text{in}}+d_{\text{out}})}$, enhancing $s_i$; and (2) a scaling factor $f_{W_s}(\tilde{s}_{i}) \in \mathbb{R}^{l_e\times 1}$, adjusting update strength across tokens.
The modified gradient decomposition $\hat{s}_{i}$ is then computed as:
\begin{gather}
\hat{s}_{i} = \sigma\left(f_{W_s}(\tilde{s}_{i})\right) \odot \left(s_i + f_{W_r}(\tilde{s}_{i})\right) \in \mathbb{R}^{l_e\times (d_{\text{in}}+d_{\text{out}})}
\end{gather}
where $\sigma$ denotes the sigmoid function and $\odot$ denotes the element-wise product, applied with broadcasting. 

The concatenated vector $\hat{s}_{i}$ is then split back into $\hat{h}_{i} \in \mathbb{R}^{l_e \times d_{\text{in}}}$ and $\hat{u}_{i} \in \mathbb{R}^{l_e \times d_{\text{out}}}$. 
Finally, the edit is executed by updating the matrices $\{W_i\}_{i=1}^n$ with the adapted gradients:
\begin{gather}
\hat{W}_i = W_i - \eta_i \hat{h}_i^\top \hat{u}_i, \quad i = 1, \ldots, n
\end{gather}
where $\eta_i$ is a trainable parameter serving as the learning rate for the edit gradients.
For a batch of edit requests, this update is applied iteratively across the batch.

\subsubsection{Training Loss}
Given a batch of requests $\mathcal{E}$ and noise $\epsilon$, we denote the IBKE-edited LLM as $f_{\hat{\theta}_{\mathcal{E}}^\epsilon}$.
Using the SG (sufficiency for generality) and IL (independence for locality) constraints—and by empirically approximating the true data distribution—we define the batch-wise losses as:
\begin{align}
\ell_{\text{SG}}(\mathcal{E}) &= - \frac{1}{|\mathcal{G}(\mathcal{E})|} \sum_{(q_g, o_g) \in \mathcal{G}(\mathcal{E})} \log f_{\hat{\theta}_{\mathcal{E}}^\epsilon}(o_g \mid q_g)\\
\ell_{\text{IL}}(\mathcal{E}) &= \frac{1}{|\mathcal{L}(\mathcal{E})|} \sum_{(q_l, o_l) \in \mathcal{L}(\mathcal{E})} \mathrm{KL} \left[ f_{\hat{\theta}_{\mathcal{E}}^{\epsilon}}(o_l \mid q_l) \,\|\, f_\theta(o_l \mid q_l) \right]
\label{eq_IL_loss}
\end{align}
where $\epsilon \sim \mathcal{N}(\mathbf{0}, \mathbf{I})$.
Here, differing from Eq.~\ref{eq_IB_obj_independence}, $\ell_{\text{IL}}(\mathcal{E})$ replaces the autoregressively generated locality output $\hat{o}_l$ with real outputs $o_l$ from $\mathcal{L}(e)$ to enable parallel token-wise loss computation.

Together with the ITM constraint, the total edit training loss is:
\begin{gather}
\ell_{\text{IBKE}} =  \ell_{\text{SG}}(\mathcal{E}) + \ell_{\text{IL}}(\mathcal{E}) + \frac{\beta}{|\mathcal{E}| n l_m d_m} \sum_{e \in \mathcal{E}} \sum_{i=1}^n \ell_{\text{ITM}_i}(e)
\end{gather}
where the denominator $|\mathcal{E}| n l_m d_m$ normalizes the ITM constraint at the element level, preventing the IB regularization term from becoming excessively large.

\section*{Data Availability}
The datasets used in this study are publicly available from the following sources: UniEdit (\url{https://huggingface.co/datasets/qizhou/UniEdit}), MQuAKE (\url{https://github.com/princeton-nlp/MQuAKE}), CounterFact (\url{https://rome.baulab.info/}), ZSRE (\url{https://github.com/eric-mitchell/mend}).

\bibliographystyle{IEEEtran}
\bibliography{main}

\vfill

\end{document}